\PassOptionsToPackage{table,xcdraw}{xcolor}
\documentclass[10pt,twocolumn,letterpaper]{article}

\usepackage{cvpr}

\usepackage[T1]{fontenc}
\usepackage[utf8]{inputenc}
\usepackage{times}
\usepackage{epsfig}
\usepackage{graphicx}
\usepackage{amsmath}
\usepackage{amsfonts}
\usepackage{amssymb}
\usepackage{array}
\usepackage{booktabs}
\usepackage{multirow}
\usepackage{xcolor}
\usepackage{pifont}
\usepackage[most]{tcolorbox}
\usepackage{microtype}

\definecolor{cvprblue}{rgb}{0.21,0.49,0.74}
\usepackage[pagebackref,breaklinks,colorlinks,allcolors=cvprblue]{hyperref}

\def\paperID{10743}
\def\confName{CVPR}
\def\confYear{2025}

\title{On Success and Simplicity: A Second Look at Transferable Vision-Language Attack Pipeline}

\author{
Yuchen Ren$^{1}$ \quad
Zhengyu Zhao$^{1}$\thanks{Corresponding Author: Zhengyu Zhao} \quad
Chenhao Lin$^{1}$ \quad
Bo Yang$^{2}$ \quad
Chao Shen$^{1}$\\
$^{1}$School of Cyber Science and Engineering, Xi'an Jiaotong University, Xi'an, China\\
$^{2}$State Key Laboratory of Mathematical Engineering and Advanced Computing,\\
Information Engineering University, Zhengzhou, China
}

\begin{document}
\maketitle

\begin{abstract}
Vision-Language Pre-training Models (VLPMs) are known to be vulnerable to adversarial attacks. Recent transferable attacks on VLPMs have followed a common pipeline with complicated loss functions or multi-stage text/image attacks. However, in this paper, we demonstrate that such a sophisticated attack pipeline can be simpler yet more successful. Specifically, we identify three previously overlooked issues caused by inappropriate cross-modal interactions and excessive operations. To address them, we propose the Simple Vision-Language Attack (SimVLA) pipeline, which observably improves transferability and efficiency. Experiments on four datasets and three downstream tasks validate the superiority of our pipeline. For instance, on Flickr30k text-image retrieval dataset, our SimVLA outperforms the SOTA baseline in R@1 transferability by 8.01\%-14.71\%, while consuming only about 35.73\% of the time and 46.26\% of the max VRAM. Overall, the superiority of our SimVLA highlights the importance of leveraging domain knowledge (e.g., our proposed cross-modal word identification), while blindly pursuing intricate operations (e.g, complex loss functions and redundant multi-stage designs) may even be harmful. We hope our SimVLA can serve as a simple yet effective backbone for future extensions. 
Code is available at \url{https://github.com/RYC-98/SimVLA}.

\end{abstract}

\begin{center}
	\small
	This article has been accepted for publication in
	\textit{IEEE Transactions on Information Forensics and Security}.
	This is the author's accepted manuscript.
	The version of record will be available at
	\href{https://doi.org/10.1109/TIFS.2026.3714129}
	{https://doi.org/10.1109/TIFS.2026.3714129}.
	
	\copyright~2026 IEEE. Personal use of this material is permitted.
	Permission from IEEE must be obtained for all other uses, in any current
	or future media, including reprinting/republishing this material for
	advertising or promotional purposes, creating new collective works, for
	resale or redistribution to servers or lists, or reuse of any copyrighted
	component of this work in other works.
\end{center}

\begin{figure}[t]
	\centering
	\includegraphics[width=0.99\columnwidth]{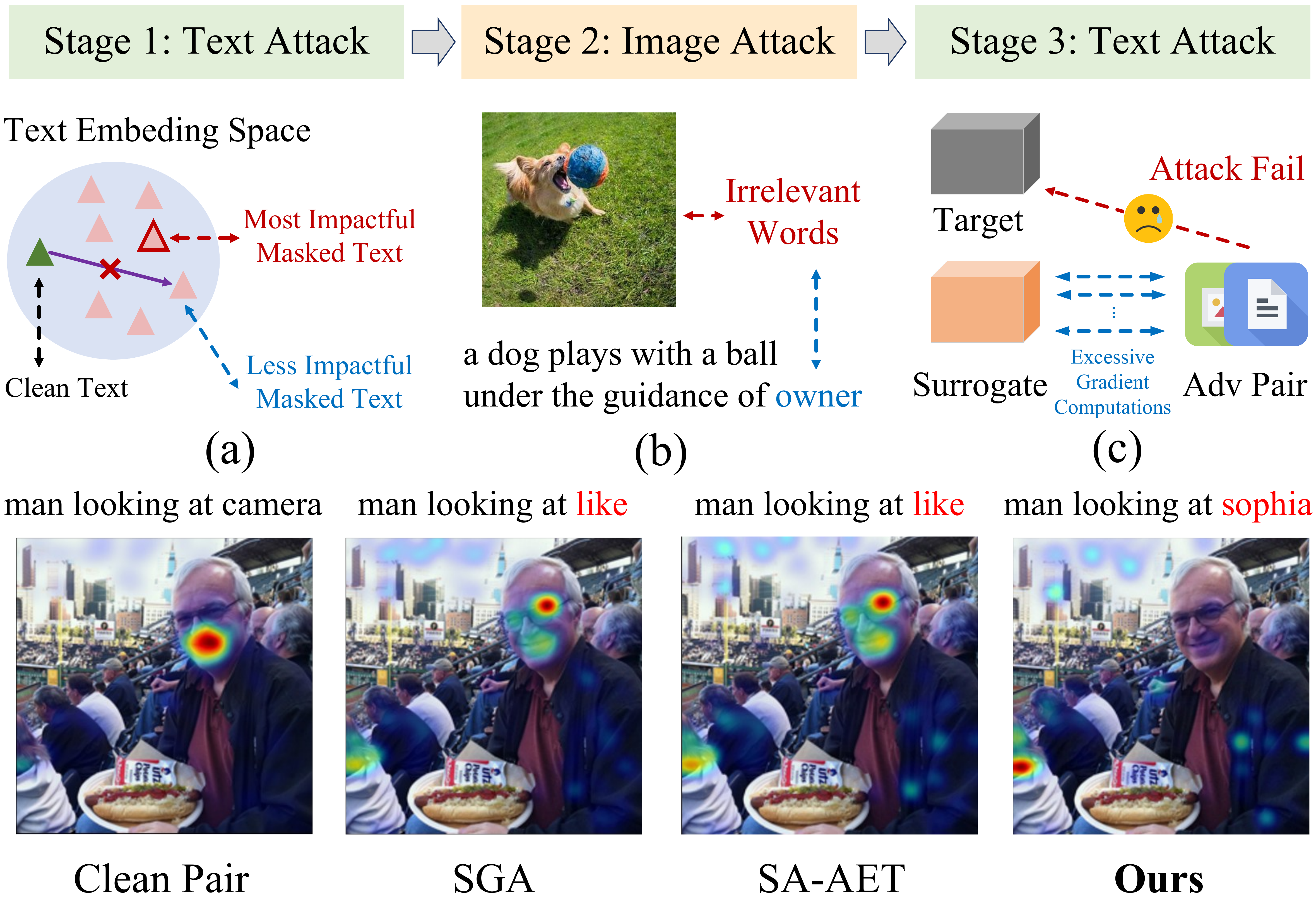}
	\caption{
		We identify three stage-specific problems in the common attack pipeline.
		\textbf{Top:} (a) Uni-modal word identification,
		(b) involving image-irrelevant words, and (c) excessive dependence on surrogate.
		\textbf{Bottom:} Our attack disrupts the image features more severely by shifting the Grad-CAM~\cite{cam} attention more than the state of the art, SGA \cite{sga} and SA-AET \cite{saaet}.
		Here, the surrogate (target) model is $\rm CLIP_{CNN}$ \cite{clip} (ALBEF \cite{albef}).}
	\label{prob}
\end{figure}
\section{Introduction}




Generating adversarial examples on vision--language pre-trained models (VLPMs) involves not a single attack 
but a pipeline of multiple attacks across different modalities. 
So far, transferable attacks on VLPMs have often drawn inspiration from other research domains \cite{contrastattack,ot,acm2024,reviewer1_advclip}, particularly contrastive learning \cite{circle,triple}, to design complex cross-modal loss functions \cite{contrastattack,reviewer1_advclip}. The core idea is generally to push adversarial samples away from benign samples under different views.
In fact, we also explored similar strategies in our preliminary experiments. For example, we constructed circle loss \cite{circle} or triplet loss \cite{triple} using the current text-image pair together with different pairs. Unfortunately, these attempts mainly improved white-box attack success rates, while offering little to no gain in black-box transferability.
In addition, many existing methods habitually follow a modified three-stage pipeline \cite{sga,sa,dra,saaet}: text attack $\rightarrow$ image attack $\rightarrow$ text attack. Specifically, Stage 1 uses clean text and clean images to generate adversarial text; Stage 2 uses adversarial text and clean images to generate adversarial images; and Stage 3 uses clean text and adversarial images to further update adversarial text.
However, the underlying mechanisms of such a sophisticated three-stage pipeline, as well as quantitative analyses validating its associated operations, have not been well justified in those works.
Blindly following these complicated designs yields minimal improvements in transferability, as they will  exacerbate overfitting of multimodal attacks.
Again, taking the three-stage pipeline as an example, the adversarial image generated in the second stage is optimized to increase its embedding distance from the adversarial text produced in the first stage. However, in the third stage, the optimization restarts from clean text and again pushes its embedding away from that of the current adversarial image. This alternating optimization can lead to over-optimization on the surrogate model: although the text--image embedding distance is enlarged on the surrogate model, it does not necessarily generalize to other target models. Prior studies have shown that improved transferability does not stem from simply maximizing the loss on the surrogate model, but rather from converging to relatively flat regions of the loss landscape \cite{rap,ncs}. Our subsequent quantitative results in \autoref{2ta} further confirm this observation.

Considering the complex nature of text-image interactions and well-recognized difficulty in attack transferability, in this paper, we revisit this common attack pipeline.
Specifically, we identify the following three problems, with one problem per stage (as illustrated in \autoref{prob}).
\noindent\textbf{Uni-modal word identification.}
In Stage 1, BERT-Attack \cite{BERTattack} is commonly adopted to generate adversarial texts. A crucial step in BERT-Attack is to identify the most impactful word in the clean text for replacement. 
However, existing identification criteria only consider text modality, overlooking that the impact should also involve the paired image, and thus making it hard to identify the truly impactful words within the text. 
This is because, in text--image retrieval attacks, the objective is to disrupt the semantic alignment between the text and its paired image, rather than the consistency within the text itself. As a result, words that appear impactful from a text-only perspective may not be the ones that most strongly influence cross-modal similarity, making image-aware identification essential.
\noindent\textbf{Involving image-irrelevant words.}
In Stage 2, the text often contains image-irrelevant information.
Therefore, directly using the raw text embeddings to guide the generation of adversarial images may provide an unnecessary optimization effort for adversarial image perturbation.
An intuitive example is an image of a dog playing with a ball, paired with the sentence ``a dog plays with a ball under the guidance of owner''. In this sentence, some words such as ``owner'' may be less semantically relevant to the visual content, because they do not explicitly appear in the image or are only partially represented, as shown in \autoref{prob} (c). When generating adversarial image perturbations, the goal is to push the text away from visually grounded concepts such as ``dog'' and ``ball,'' rather than paying efforts on unrelated words like ``owner.'' Therefore, relying on the raw text to guide adversarial generation can be inaccurate.
\noindent\textbf{Excessive dependence on surrogate.}  
In Stage 3, adversarial texts are generated again.
However, such repeating of the text attack may cause excessive gradient computations on the surrogate model, resulting in potential overfitting to the information of the surrogate model.

To address the above three problems, we propose the Simple Vision-Language Attack (SimVLA) pipeline, which systematically refines the common attack pipeline by enriching cross-modal interactions and removing unnecessary operations.
Concretely, in Stage 1, we identify the most impactful word based on the cross-modal information rather than the text.
In Stage 2, before generating the adversarial images, we first remove the most irrelevant words from the supervising text.
Finally, based on qualitative and quantitative analysis, we remove Stage 3, which repeats the text attack, to prevent potential overfitting to the surrogate model.
\autoref{prob} shows that compared to existing attacks, our SimVLA can better disrupt the text-image alignment by shifting the original image attention more.

Our main contributions are as follows:
\begin{itemize}
	\item 
	We revisit the common three-stage pipeline of transferable attacks on VLPMs, identifying one problem per stage that is caused by the lack of correct cross-modal interactions or excessive unnecessary operations.
	
	\item 
	We propose the Simple Vision-Language Attack (SimVLA) pipeline, consisting of cross-modal word identification, text semantic abstraction, and stage deletion, to systematically address the three problems.

	
	
	
	\item
	Extensive experiments demonstrate the superiority of SimVLA in both transferability and efficiency. For example,  when compared to SOTA baseline in retrieval tasks on Flickr30k, SimVLA improves transferability by  8.01\%-14.71\%, while consuming only about 35.73\% of the time and 46.26\% of the max VRAM.
	

	
	

	
\end{itemize}

\section{Related Work}

\subsection{Transferable Adversarial Attacks on VLPMs}

Beyond attacking only one modality, 
Separate Unimodal Attack (Sep-Attack) \cite{coattack} sequentially attacks both modalities of VLPMs in white-box settings.
Collaborative Multimodal Adversarial Attack (Co-Attack) \cite{coattack} improves Sep-Attack by introducing a cross-modal interaction loss function to generate adversarial text-image pairs. 
VLATTACK \cite{reviewer1_vlaattack} tackles black-box attacks on fine-tuned models by jointly crafting image and text perturbations through block-wise similarity disruption and iterative cross-modal optimization. 
AdvCLIP \cite{reviewer1_advclip} constructs a universal adversarial patch by modeling cross-modal topology and generating topology-deviation perturbations, enabling non-targeted attacks on diverse tasks inheriting the same pre-trained encoder.
ETU \cite{reviewer1_univlp} generates effective and transferable universal adversarial perturbations by exploiting cross-modal interactions and balancing global--local utilities. It further employs ScMix, a combination of self-mix and cross-mix transformations, to strengthen transferability.
SceneTAP \cite{reviewer3_scenetap} uses LLM chain-of-thought reasoning to perform scene understanding, adversarial planning, and seamless integration, with attacks executed by a scene-coherent TextDiffuser.
VLP-Attack \cite{contrastattack} leverages text-image and intra-modal contrastive losses to jointly generate transferable adversarial texts and images.
OT-Attack \cite{ot} leverages optimal transport to align cross-modal features and improve adversarial transferability across vision-language models.
GLEAM \cite{gleam} enhances transferable attacks through global-local transformations that jointly capture holistic semantics and fine-grained regional perturbations.

In this work, we revisit the three-stage attack pipelines, which are commonly used in transferable attacks on VLPMs. Given that this three-stage optimization paradigm has been consistently adopted, our analysis aims to provide clearer insights and avoid potentially misleading future research.
Set-Level Guidance Attack (SGA) \cite{sga} proposes a three-stage attack pipeline, which is also followed by Diverse Region Attack (DRA) \cite{dra}.
Specifically, SGA constructs loss functions for set-level text-image pairs by applying scale transformation and Gaussian noise to images.
Self-Augment-based Transfer Attack (SA-Attack) \cite{sa} improves adversarial transferability by applying modality-specific data augmentations during the joint generation of adversarial images and texts.
Diverse Region Attack (DRA) \cite{dra} follows the three-stage attack pipeline and further incorporates several additional techniques, including intersection region diversification, text-guided augmentation selection, and an intersection region--based multi-term loss function to guide adversarial text generation.
Semantically Aligned Adversarial Evolution Triangle method (SA-AET) \cite{saaet} is an extension of DRA and has become the current SOTA attack method on VLPMs. It enhances adversarial diversity by sampling from adversarial evolution triangles constructed from clean, historical, and current examples. It further improves transferability by generating adversarial examples in a semantic image--text contrastive subspace that reduces redundant feature dimensions and mitigates model-dependent similarity distortions.
Unfortunately, although SA-AET improves transferability, it still relies on a three-stage pipeline and introduces even more complex optimization procedures, which further increase the computational burden on the attacker (see the efficiency comparison in \autoref{cost}).

To address these problems, we propose to enrich cross-modal interactions by leveraging the embedding center derived from a group of sampled images to identify the most influential words. Moreover, we remove unnecessary operations by directly eliminating the third stage, thereby substantially improving both transferability and efficiency.

\begin{figure*}[t]
	\centering
	\includegraphics[width=2.1\columnwidth]{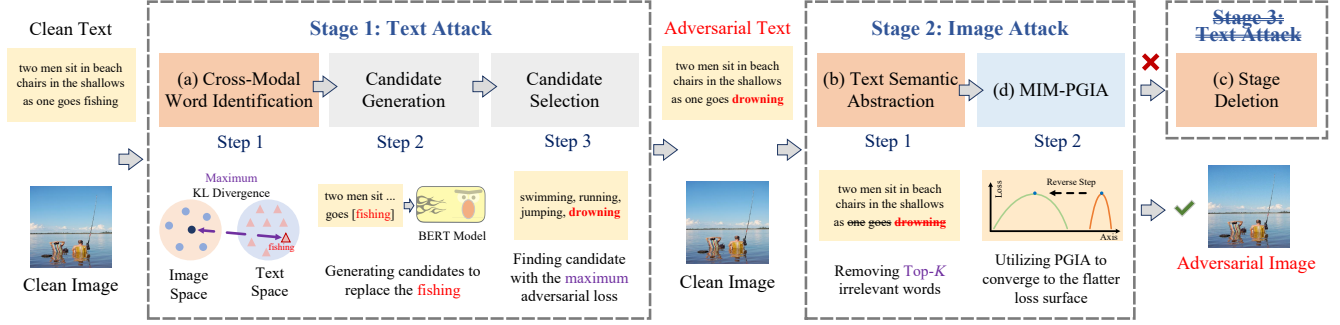} 
	\caption{
		The overview of our Simple Vision-Language Attack (SimVLA) pipeline. Our contributions focus on (a) cross-modal word identification (\textbf{Stage 1}, \textbf{Step 1}) to effectively disrupt text-image alignment; (b) text semantic abstraction (\textbf{Stage 2}, \textbf{Step 1}) to filter out irrelevant words in the adversarial text; (c) stage deletion (\textbf{Stage 3}) to mitigate potential overfitting.
        Besides,
		(d) Momentum-based Previous Gradient Inversion
		Approximation (MIM-PGIA) \cite{ncs} is also adopted in Stage 2 to further improve transferability based on loss flatness.
	}
	\label{overview}
\end{figure*}

\section{Methodology}
\subsection{Preliminary}
Transferable attacks on VLPMs aim to generate adversarial text-image pair $({x_{adv}},{t_{adv}})$ on the surrogate VLPM $f$ with a clean input pair $({x},{t})$~\cite{sga,dra}. 
The output embeddings\footnote{For fused VLPMs, such as ALBEF \cite{albef} and TCL \cite{tcl}, the output embedding refers to the CLS token from the final layer. For aligned VLPMs like CLIP \cite{clip}, the output embedding is the output representation itself.} of the image encoder and text encoder are denoted as $f_I(x)$ and $f_T(t)$, respectively. 
The generation of adversarial text-image pair can be formulated as: 
\begin{equation}
	\label{total_goal}
	({x_{adv}, {t_{adv}}}) = \mathop {\arg \max }\limits_{{\rm{ }}x \in B(x,\epsilon_{x}), t \in B(t,\epsilon_{t})}  L({f_I}(x),{f_T}(t)),
\end{equation}
where $L$ is the loss functions, $B$ is the feasible region of adversarial text-image pair, $\epsilon_{t}$ is the maximum number of words that can be replaced in the clean text $t$ (any word substitution is allowed as long as the total replaced number remains within $\epsilon_{t}$), and $\epsilon_{x}$ is the image perturbation bound imposed on the clean image $x$. 

The adversarial text and image are often optimized alternately, following a three-stage pipeline originally introduced by SGA \cite{sga}.
All these three stages aim to push away the paired text and image but differ in the text or image versions they use.
Specifically, Stage 1 (BERT-Attack \cite{BERTattack}) uses clean text and clean image to generate adversarial text, Stage 2 (PGD \cite{pgd}) uses adversarial text and clean image to generate adversarial image, and Stage 3 (BERT-Attack \cite{BERTattack}) uses clean text and the previously generated adversarial image to generate the final adversarial text.
These attacks are characterized by improper modality interactions or complex loss functions, which leads to overfitting in white-box models. They typically require significant computational time but yield limited improvement.

To address previous limits, we thereby introduce our Simple Vision-Language Attack (SimVLA) pipeline, consisting of three strategies: cross-modal word identification (Stage 1), text semantic abstraction (Stage 2), and stage deletion (Stage 3), along with the optimization details of Stage 1 and Stage 2.
\autoref{overview} provides an overview of our SimVLA. 
Importantly, the core contributions of SimVLA are the three stage-wise refinements: cross-modal word identification, text semantic abstraction, and stage deletion. MIM-PGIA is an auxiliary off-the-shelf optimizer adopted only for adversarial image optimization in Stage 2, rather than a component proposed in this work.

\subsection{Text Attack (Stage 1)}
\subsubsection{Cross-Modal Word Identification}

Text attacks in existing studies (SGA, DRA, SA-AET, etc.) are based on BERT-Attack, which includes three standard steps: 1) Identifying the most influential word; 2) Candidate generation; 3) Word replacement. Details are as follows:
\begin{enumerate}
	\item Identifying the word that has the greatest impact on the overall semantics of the text. This is achieved by generating a masked text set $S_{mask}$, with a length equal to the number of words in the text $t$. Each masked text $t'$ in $S_{mask}$ has a word at a specific position replaced with `[UNK]'. If a particular masked text $t'$ causes the largest change in KL divergence compared to the clean text embedding ${f_T}(t)$, then the word masked in $t'$ is identified as the most impactful word $w^*$:
	\begin{equation}
		\label{kltxt}
		{w^*} = \mathop {\arg \max }\limits_{t' \in {S_\mathrm{mask}}} KL({f_T}(t)\parallel {f_T}(t')).
	\end{equation}
	
	\item Using the BERT model \cite{BERT} to generate a group of semantically related candidate words for the identified word in Step 1.
	
	\item Optimizing the adversarial text by iteratively replacing the identified word with the candidate word that maximizes the adversarial loss.
\end{enumerate}

Previous work has primarily focused on Step 3, combining various loss functions and extensive data augmentations.
For example, SA-Attack \cite{sa} employs more than 4 types of data augmentations and a two-term loss function: one term separates the optimized text from the clean image, while the other separates it from the combined clean and adversarially augmented images in the embedding space. 
DRA \cite{dra} and SA-AET \cite{saaet} further formulate adversarial text generation with a three-term objective, where the loss jointly considers the clean image, the previous adversarial image, and the current adversarial image (i.e., an adversarial triangle) to guide optimization. 
VLP-Attack \cite{contrastattack} adopts an even more obscure loss function, constructing a five-term loss function composed of adversarial loss, cosine similarity loss, likelihood loss, cross-modal loss, and intra-modal loss.
However, whether these complex loss terms are intrinsically consistent and whether each component is truly necessary remain insufficiently analyzed.

Instead of focusing on stacking data augmentations and increasingly complex loss functions, we notice that in Step 1, existing work has blindly followed the original BERT-Attack, which only considers text modality. 
This is problematic because, in VLPM attacks, the most influential word should depend on the paired image rather than the text itself, due to the inherent nature of multimodal tasks.
Motivated by this insight, we propose cross-modal word identification, which
substitutes the clean text embedding with the image embedding to compute the KL divergence\footnote{KL divergence is commonly employed in VLPM attacks \cite{sa,sga,dra,coattack} to assess the difference in cross-modal embeddings.} against each masked texts embeddings. 
To represent image features, we utilize the embedding center derived from a group of sampled images in the neighborhood, and compute the KL divergence between this image embedding center and the embeddings of each masked text.

To obtain the sampled images in the neighborhood, we 
add random noise $\delta_{i}$ drawn from the uniform distribution $U[ - \beta  \cdot \epsilon_x,\beta  \cdot \epsilon_x]$ to the clean image $x$, as in previous work \cite{vt,gra}. 
Then, the image embedding center $f_\mathrm{center}(x)$  is computed by:
\begin{equation}
	\label{center}
	{f_\mathrm{{center}}}(x) = \frac{1}{{{N_I}}}\sum\limits_{i = 1}^{{N_I}} {{f_I}(x + {\delta _i})}, 
\end{equation}
where $N_I$ is the sample number and $\beta$ represents the boundary factor.
Process of identifying the most influential word can be written as: 
\begin{equation}
	\label{klimg}
	{w^*} = \mathop {\arg \max }\limits_{t' \in {S_\mathrm{mask}}} KL({f_\mathrm{{center}}}(x)\parallel {f_T}(t')).
\end{equation}

\subsubsection{Optimization of Adversarial Text}
Considering that Step 2 and Step 3 of Stage 1 have been detailed in previous work \cite{coattack,sa,sga,dra}, and they are not our main contribution, we will directly present the final optimized loss function $ L(t)$ for text attack here:

\begin{equation}
	\label{text_loss}
	L(t) =  - \frac{{{f_\mathrm{{center}}}(x) \cdot {f_T}(t)}}{{{{\left\| {f_\mathrm{{center}}(x)} \right\|}_2} \cdot {{\left\| {{f_T}(t)} \right\|}_2}}}.
\end{equation}

Solving Equation \ref{text_loss} is a fixed and discrete process, typically involving searching for the text from the given vocabulary that maximizes this loss as the final adversarial text. Details can be found in previous works \cite{BERTattack,coattack,sa,sga,dra}.

\subsection{Image Attack (Stage 2)}
\subsubsection{Text Semantic Abstraction}
In the image attack stage, the embeddings of the adversarial text $t_{adv}$, generated during the text attack stage, are used as targets to guide the generation of adversarial images. However, $t_{adv}$ often includes words irrelevant to the paired image content, which may introduce conflicting information and hinder the effective disruption of text-image alignment. Accordingly, it is necessary to remove such irrelevant words from $t_{adv}$ before proceeding with the image attack stage.

Therefore, we propose text semantic abstraction in the Step 1 of Stage 2, which is conceptually similar to the process of cross-modal word identification in Stage 1.
It begins by masking each word in $t_{adv}$, followed by assigning an importance rank to all the words in the text, using KL divergence.
The KL divergence is computed between the embedding of each masked adversarial text and 
image embedding center $f_\mathrm{center}(x)$, like Equation \ref{center}.
The key difference here is that we need to identify the Top-$K$ irrelevant words in the embedding space, rather than the most influential ones as before.
Then, we remove the Top-$K$ irrelevant words from the adversarial text, obtaining the modified adversarial text, denoted as $t_{adv}^R$, for the subsequent construction of the loss function.

To conclude, the purpose of text semantic abstraction is to avoid allocating unnecessary effort to semantically irrelevant content during image perturbation generation, which may otherwise degrade attack performance in text--image retrieval tasks.
Appendix \ref{ap:ssec:delete_stat} also reports the proportion of overlapping deleted words across different surrogate models under the Top-1 settings. The results show that, compared with random selection, our method is able to identify words that are irrelevant to the image across different models.

\subsubsection{Optimization of Adversarial Image}

We provide the loss function $ L(x)$ for image attack, i.e., the Step 2 of Stage 2. Here, we also simply adopt the negative cosine similarity as the final loss: 
\begin{equation}
	\label{one_image_loss_main}
	L(x) = - \sum\limits_{j = 1}^{{N_T}} {\frac{{{f_I}(x) \cdot {f_T}({{({t_j})_{adv}^R}})}}{{{{\left\| {{f_I}(x)} \right\|}_2} \cdot {{\left\| {{f_T}({{({t_j})_{adv}^R}})} \right\|}_2}}}},
\end{equation} 
where $N_T$ denotes the number of ground-truth texts that are paired with the clean image $x$ in the dataset, and $t_j$ is the $j$-th ground-truth text of clean image $x$.
To solve Equation \ref{one_image_loss_main}, we adopt the momentum-based Previous Gradient Inversion Approximation (MIM-PGIA) \cite{ncs}, which directs adversarial images to converge towards a flatter loss surface, thereby enhancing transferability \cite{rap}.
Note that MIM-PGIA serves only as an auxiliary optimizer and does not alter the three new refinements of SimVLA. It can be replaced with the standard PGD optimizer while retaining the same attack pipeline. We separately isolate its contribution from those of the three proposed components through controlled ablation experiments in \autoref{abl}. 
See more technical details of MIM-PGIA in Appendix \ref{MIM-PGIA}.

\subsection{Stage Deletion (Stage 3)}

This stage is widely adopted by previous work based on the assumption that it can push adversarial text further away from the adversarial image, and so increasing the adversarial loss \cite{sga,sa,dra}.
Although this assumption is true for white-box attacks, it may lead to redundant gradient interactions with the surrogate model, increasing the risk of overfitting to the surrogate. 
To validate this, we compare the white-box text-image average similarity during the optimization process on the surrogate model, and the black-box attack success rates (transferability) of text-image retrieval task.

\begin{table}[!htb] 
	\caption{Effect of removing the second text attack stage (Stage 3). Transferability is based on R@k in text retrieval (TR) and image retrieval (IR). The dataset is Flickr30k and the surrogate (target) model is ALBEF ($\rm CLIP_{ViT}$).
}
\label{2ta}
	\centering
	\resizebox{\linewidth}{!}{
		
		\begin{tabular}{cccccc}
			\toprule[1pt]
			\multirow{2}{*}{Attack} & Similarity & \multicolumn{2}{c}{TR (Black-box)} & \multicolumn{2}{c}{IR} (Black-box) \\
			& (White-box) & R@1 & R@10 & R@1 & R@10 \\ \toprule[1pt]
			SGA & \textbf{-0.093} & 32.39 & 8.64 & 44.07 & 21.00 \\
			SGA w/o Stage 3 & -0.083 & \textbf{35.71} & \textbf{8.84} & \textbf{48.32} & \textbf{22.94} \\ \toprule[1pt]
		\end{tabular}
	}
\end{table}

As can be seen from \autoref{2ta}, SGA, retaining the second text attack (Stage 3), results in lower average similarity on the white-box surrogate model, which indicates the embeddings of adversarial text-image pairs are more irrelevant.
However, the attack success rate of SGA is lower on the target model, indicating the existence of overfitting.
To alleviate this overfitting, we simply delete Stage 3 from the common attack pipeline, which also improves the attack efficiency.

\begin{table*}[!htb]
	\caption{
	Cross-model attack success rates (\%) regarding \textbf{R@1} in Text Retrieval (TR) and Image Retrieval (IR) on Flickr30k and MSCOCO. 
	Note that the white-box success (in gray) is not the focus of transferable attacks.}
\label{body_r1}
	\centering
	\Large
	\resizebox{\linewidth}{!}{
\begin{tabular}{cccccccccccccccccc}
\toprule[1pt]
 &  & \multicolumn{4}{c}{ALBEF} & \multicolumn{4}{c}{TCL} & \multicolumn{4}{c}{$\rm CLIP_{ViT}$} & \multicolumn{4}{c}{$\rm CLIP_{CNN}$} \\
 &  & \multicolumn{2}{c}{Flickr30k} & \multicolumn{2}{c}{MSCOCO} & \multicolumn{2}{c}{Flickr30k} & \multicolumn{2}{c}{MSCOCO} & \multicolumn{2}{c}{Flickr30k} & \multicolumn{2}{c}{MSCOCO} & \multicolumn{2}{c}{Flickr30k} & \multicolumn{2}{c}{MSCOCO} \\ \cmidrule(lr){3-6} \cmidrule(lr){7-10} \cmidrule(lr){11-14} \cmidrule(lr){15-18}
\multirow{-3}{*}{Model} & \multirow{-3}{*}{Attack} & TR & IR & TR & IR & TR & IR & TR & IR & TR & IR & TR & IR & TR & IR & TR & IR \\ \toprule[1pt]
 & PGD & \cellcolor[HTML]{EFEFEF}57.04 & \cellcolor[HTML]{EFEFEF}62.09 & \cellcolor[HTML]{EFEFEF}67.70 & \cellcolor[HTML]{EFEFEF}73.07 & 3.69 & 8.12 & 10.19 & 14.03 & 8.34 & 13.14 & 14.19 & 22.59 & 9.96 & 15.23 & 17.29 & 23.76 \\
 & BERT-Attack & \cellcolor[HTML]{EFEFEF}11.57 & \cellcolor[HTML]{EFEFEF}27.48 & \cellcolor[HTML]{EFEFEF}35.86 & \cellcolor[HTML]{EFEFEF}47.39 & 12.75 & 28.07 & 35.93 & 44.84 & 29.20 & 43.20 & 53.19 & 63.43 & 32.57 & 46.11 & 56.19 & 65.04 \\
 & Sep-Attack & \cellcolor[HTML]{EFEFEF}69.45 & \cellcolor[HTML]{EFEFEF}76.43 & \cellcolor[HTML]{EFEFEF}84.02 & \cellcolor[HTML]{EFEFEF}87.11 & 18.55 & 33.48 & 41.64 & 51.30 & 29.82 & 45.23 & 54.37 & 63.88 & 32.24 & 46.11 & 56.85 & 65.17 \\
 & Co-Attack & \cellcolor[HTML]{EFEFEF}69.86 & \cellcolor[HTML]{EFEFEF}76.45 & \cellcolor[HTML]{EFEFEF}84.07 & \cellcolor[HTML]{EFEFEF}87.45 & 18.65 & 33.64 & 42.22 & 51.19 & 29.57 & 44.94 & 53.87 & 64.10 & 32.44 & 46.28 & 57.17 & 65.03 \\
 & SGA & \cellcolor[HTML]{EFEFEF}96.98 & \cellcolor[HTML]{EFEFEF}97.22 & \cellcolor[HTML]{EFEFEF}96.62 & \cellcolor[HTML]{EFEFEF}96.86 & 45.84 & 55.43 & 58.76 & 65.48 & 32.39 & 44.07 & 58.11 & 65.44 & 35.12 & 46.79 & 58.77 & 66.70 \\
 & DRA & \cellcolor[HTML]{EFEFEF}96.30 & \cellcolor[HTML]{EFEFEF}96.61 & \cellcolor[HTML]{EFEFEF}96.42 & \cellcolor[HTML]{EFEFEF}96.59 & 48.16 & 57.40 & 59.95 & 66.64 & 36.69 & 47.97 & 59.79 & 67.27 & 39.59 & 49.78 & 61.30 & 68.17 \\
 & SA-AET & \cellcolor[HTML]{EFEFEF}97.08 & \cellcolor[HTML]{EFEFEF}97.40 & \cellcolor[HTML]{EFEFEF}98.61 & \cellcolor[HTML]{EFEFEF}98.50 & 54.69 & 64.19 & 70.56 & 74.86 & 37.81 & 50.48 & 61.50 & 68.47 & 45.40 & 54.57 & 65.59 & 73.70 \\
\multirow{-8}{*}{ALBEF} & SimVLA & \cellcolor[HTML]{EFEFEF}94.79 & \cellcolor[HTML]{EFEFEF}95.09 & \cellcolor[HTML]{EFEFEF}95.88 & \cellcolor[HTML]{EFEFEF}96.14 & \textbf{63.65} & \textbf{72.88} & \textbf{76.30} & \textbf{80.57} & \textbf{52.52} & \textbf{60.24} & \textbf{71.12} & \textbf{76.38} & \textbf{56.69} & \textbf{65.88} & \textbf{75.24} & \textbf{81.29} \\ \hline
 & PGD & 7.82 & 13.26 & 13.87 & 19.21 & \cellcolor[HTML]{EFEFEF}81.98 & \cellcolor[HTML]{EFEFEF}84.33 & \cellcolor[HTML]{EFEFEF}83.20 & \cellcolor[HTML]{EFEFEF}85.39 & 7.24 & 13.53 & 14.35 & 22.71 & 10.34 & 15.44 & 17.00 & 23.55 \\
 & BERT-Attack & 11.89 & 26.85 & 36.01 & 47.34 & \cellcolor[HTML]{EFEFEF}14.54 & \cellcolor[HTML]{EFEFEF}29.17 & \cellcolor[HTML]{EFEFEF}37.78 & \cellcolor[HTML]{EFEFEF}47.98 & 29.57 & 44.49 & 55.25 & 63.80 & 33.46 & 46.07 & 57.09 & 65.95 \\
 & Sep-Attack & 23.36 & 37.49 & 44.93 & 56.38 & \cellcolor[HTML]{EFEFEF}85.46 & \cellcolor[HTML]{EFEFEF}89.50 & \cellcolor[HTML]{EFEFEF}91.40 & \cellcolor[HTML]{EFEFEF}92.70 & 29.94 & 44.36 & 54.79 & 64.51 & 33.46 & 45.52 & 58.28 & 65.96 \\
 & Co-Attack & 22.94 & 38.17 & 45.11 & 56.64 & \cellcolor[HTML]{EFEFEF}87.88 & \cellcolor[HTML]{EFEFEF}89.33 & \cellcolor[HTML]{EFEFEF}90.50 & \cellcolor[HTML]{EFEFEF}92.67 & 31.04 & 45.07 & 55.36 & 64.29 & 33.84 & 45.90 & 58.40 & 66.28 \\
 & SGA & 48.91 & 59.85 & 65.22 & 73.18 & \cellcolor[HTML]{EFEFEF}98.52 & \cellcolor[HTML]{EFEFEF}98.71 & \cellcolor[HTML]{EFEFEF}98.99 & \cellcolor[HTML]{EFEFEF}99.13 & 32.76 & 45.10 & 57.04 & 63.91 & 37.42 & 47.92 & 58.93 & 65.49 \\
 & DRA & 50.68 & 61.62 & 68.11 & 75.73 & \cellcolor[HTML]{EFEFEF}98.10 & \cellcolor[HTML]{EFEFEF}98.31 & \cellcolor[HTML]{EFEFEF}98.86 & \cellcolor[HTML]{EFEFEF}98.98 & 36.56 & 48.07 & 61.12 & 67.44 & 42.15 & 51.08 & 62.65 & 69.22 \\
 & SA-AET & 55.53 & 65.57 & 71.67 & 78.60 & \cellcolor[HTML]{EFEFEF}98.02 & \cellcolor[HTML]{EFEFEF}98.18 & \cellcolor[HTML]{EFEFEF}98.22 & \cellcolor[HTML]{EFEFEF}98.49 & 36.20 & 48.97 & 62.00 & 68.52 & 45.15 & 54.70 & 66.50 & 74.44 \\
\multirow{-8}{*}{TCL} & SimVLA & \textbf{63.86} & \textbf{74.17} & \textbf{79.89} & \textbf{85.17} & \cellcolor[HTML]{EFEFEF}97.50 & \cellcolor[HTML]{EFEFEF}97.59 & \cellcolor[HTML]{EFEFEF}98.04 & \cellcolor[HTML]{EFEFEF}98.39 & \textbf{47.61} & \textbf{60.15} & \textbf{71.50} & \textbf{76.47} & \textbf{55.83} & \textbf{65.43} & \textbf{75.73} & \textbf{81.64} \\ \hline
 & PGD & 1.98 & 4.91 & 7.24 & 11.03 & 3.90 & 6.83 & 8.57 & 12.25 & \cellcolor[HTML]{EFEFEF}69.57 & \cellcolor[HTML]{EFEFEF}74.94 & \cellcolor[HTML]{EFEFEF}83.86 & \cellcolor[HTML]{EFEFEF}86.62 & 5.75 & 8.34 & 9.11 & 13.55 \\
 & BERT-Attack & 9.18 & 22.71 & 28.31 & 40.95 & 9.69 & 24.12 & 29.21 & 41.01 & \cellcolor[HTML]{EFEFEF}28.34 & \cellcolor[HTML]{EFEFEF}39.01 & \cellcolor[HTML]{EFEFEF}55.13 & \cellcolor[HTML]{EFEFEF}57.79 & 30.40 & 37.32 & 52.31 & 58.58 \\
 & Sep-Attack & 9.70 & 22.80 & 28.41 & 41.22 & 10.64 & 24.74 & 30.53 & 42.09 & \cellcolor[HTML]{EFEFEF}78.65 & \cellcolor[HTML]{EFEFEF}84.50 & \cellcolor[HTML]{EFEFEF}91.68 & \cellcolor[HTML]{EFEFEF}94.02 & 30.91 & 39.35 & 53.62 & 60.49 \\
 & Co-Attack & 10.32 & 24.23 & 29.96 & 42.39 & 11.28 & 25.76 & 31.14 & 43.34 & \cellcolor[HTML]{EFEFEF}91.66 & \cellcolor[HTML]{EFEFEF}94.78 & \cellcolor[HTML]{EFEFEF}97.86 & \cellcolor[HTML]{EFEFEF}98.22 & 32.95 & 41.51 & 55.33 & 62.25 \\
 & SGA & 12.41 & 27.46 & 33.38 & 44.59 & 14.33 & 30.07 & 35.34 & 45.44 & \cellcolor[HTML]{EFEFEF}99.14 & \cellcolor[HTML]{EFEFEF}99.07 & \cellcolor[HTML]{EFEFEF}99.77 & \cellcolor[HTML]{EFEFEF}99.79 & 38.44 & 46.72 & 58.52 & 65.71 \\
 & DRA & 13.24 & 29.42 & 35.99 & 47.69 & 14.33 & 30.88 & 36.93 & 48.57 & \cellcolor[HTML]{EFEFEF}99.02 & \cellcolor[HTML]{EFEFEF}99.00 & \cellcolor[HTML]{EFEFEF}99.77 & \cellcolor[HTML]{EFEFEF}99.73 & 44.19 & 50.53 & 64.41 & 70.20 \\
 & SA-AET & 13.87 & 30.14 & 36.79 & 48.85 & 15.17 & 30.88 & 36.83 & 49.48 & \cellcolor[HTML]{EFEFEF}98.65 & \cellcolor[HTML]{EFEFEF}99.03 & \cellcolor[HTML]{EFEFEF}99.24 & \cellcolor[HTML]{EFEFEF}99.81 & 45.85 & 50.81 & 65.63 & 70.51 \\
\multirow{-8}{*}{$\rm CLIP_{ViT}$} & SimVLA & \textbf{22.00} & \textbf{42.40} & \textbf{47.77} & \textbf{60.76} & \textbf{23.18} & \textbf{43.48} & \textbf{49.23} & \textbf{62.30} & \cellcolor[HTML]{EFEFEF}95.21 & \cellcolor[HTML]{EFEFEF}96.91 & \cellcolor[HTML]{EFEFEF}98.82 & \cellcolor[HTML]{EFEFEF}99.25 & \textbf{55.56} & \textbf{64.12} & \textbf{77.65} & \textbf{81.38} \\ \hline
 & PGD & 1.98 & 4.89 & 6.78 & 10.31 & 3.79 & 6.57 & 8.25 & 11.50 & 0.74 & 6.70 & 5.57 & 10.26 & \cellcolor[HTML]{EFEFEF}84.16 & \cellcolor[HTML]{EFEFEF}90.15 & \cellcolor[HTML]{EFEFEF}89.25 & \cellcolor[HTML]{EFEFEF}74.74 \\
 & BERT-Attack & 9.38 & 23.32 & 28.67 & 41.12 & 11.70 & 24.38 & 30.08 & 41.21 & 27.12 & 37.56 & 52.31 & 56.68 & \cellcolor[HTML]{EFEFEF}30.40 & \cellcolor[HTML]{EFEFEF}39.97 & \cellcolor[HTML]{EFEFEF}54.56 & \cellcolor[HTML]{EFEFEF}60.74 \\
 & Sep-Attack & 8.76 & 23.34 & 29.11 & 41.21 & 11.38 & 25.05 & 30.56 & 41.66 & 27.98 & 39.79 & 52.69 & 59.00 & \cellcolor[HTML]{EFEFEF}91.70 & \cellcolor[HTML]{EFEFEF}94.72 & \cellcolor[HTML]{EFEFEF}95.50 & \cellcolor[HTML]{EFEFEF}97.47 \\
 & Co-Attack & 8.86 & 23.55 & 29.31 & 41.57 & 11.70 & 25.12 & 30.71 & 42.00 & 27.61 & 39.85 & 53.30 & 58.89 & \cellcolor[HTML]{EFEFEF}95.02 & \cellcolor[HTML]{EFEFEF}96.33 & \cellcolor[HTML]{EFEFEF}96.57 & \cellcolor[HTML]{EFEFEF}98.39 \\
 & SGA & 11.37 & 24.91 & 31.58 & 42.50 & 14.33 & 27.40 & 32.59 & 44.19 & 31.53 & 41.98 & 57.23 & 60.55 & \cellcolor[HTML]{EFEFEF}99.36 & \cellcolor[HTML]{EFEFEF}98.42 & \cellcolor[HTML]{EFEFEF}99.63 & \cellcolor[HTML]{EFEFEF}99.82 \\
 & DRA & 12.62 & 26.64 & 32.77 & 44.95 & 14.33 & 29.12 & 33.84 & 46.48 & 33.13 & 46.13 & 60.09 & 64.57 & \cellcolor[HTML]{EFEFEF}99.36 & \cellcolor[HTML]{EFEFEF}99.25 & \cellcolor[HTML]{EFEFEF}99.39 & \cellcolor[HTML]{EFEFEF}99.53 \\
 & SA-AET & 12.41 & 27.15 & 33.40 & 45.23 & 14.96 & 29.31 & 34.21 & 46.60 & 37.55 & 47.62 & 62.20 & 68.41 & \cellcolor[HTML]{EFEFEF}99.87 & \cellcolor[HTML]{EFEFEF}99.52 & \cellcolor[HTML]{EFEFEF}99.81 & \cellcolor[HTML]{EFEFEF}99.69 \\
\multirow{-8}{*}{$\rm CLIP_{CNN}$} & SimVLA & \textbf{22.10} & \textbf{41.40} & \textbf{44.99} & \textbf{57.72} & \textbf{25.92} & \textbf{44.14} & \textbf{47.35} & \textbf{60.43} & \textbf{50.60} & \textbf{61.68} & \textbf{74.95} & \textbf{79.05} & \cellcolor[HTML]{EFEFEF}97.96 & \cellcolor[HTML]{EFEFEF}99.07 & \cellcolor[HTML]{EFEFEF}99.14 & \cellcolor[HTML]{EFEFEF}99.62 \\ \toprule[1pt]
\end{tabular}
}
\end{table*}

\begin{table*}[!htb]
	\caption{
	Cross-model attack success rates (\%) regarding \textbf{R@10} in Text Retrieval (TR) and Image Retrieval (IR) on Flickr30k and MSCOCO. Note that the white-box success (in gray) is not the focus of transferable attacks.}
\label{body_r10}
	\centering
	\Large
	\resizebox{\linewidth}{!}{
\begin{tabular}{cccccccccccccccccc}
\toprule[1pt]
 &  & \multicolumn{4}{c}{ALBEF} & \multicolumn{4}{c}{TCL} & \multicolumn{4}{c}{$\rm CLIP_{ViT}$} & \multicolumn{4}{c}{$\rm CLIP_{CNN}$} \\
 &  & \multicolumn{2}{c}{Flickr30k} & \multicolumn{2}{c}{MSCOCO} & \multicolumn{2}{c}{Flickr30k} & \multicolumn{2}{c}{MSCOCO} & \multicolumn{2}{c}{Flickr30k} & \multicolumn{2}{c}{MSCOCO} & \multicolumn{2}{c}{Flickr30k} & \multicolumn{2}{c}{MSCOCO} \\ \cmidrule(lr){3-6} \cmidrule(lr){7-10} \cmidrule(lr){11-14} \cmidrule(lr){15-18}
\multirow{-3}{*}{Model} & \multirow{-3}{*}{Attack} & TR & IR & TR & IR & TR & IR & TR & IR & TR & IR & TR & IR & TR & IR & TR & IR \\ \toprule[1pt]
 & PGD & \cellcolor[HTML]{EFEFEF}34.30 & \cellcolor[HTML]{EFEFEF}41.25 & \cellcolor[HTML]{EFEFEF}46.46 & \cellcolor[HTML]{EFEFEF}53.60 & 0.60 & 1.60 & 2.52 & 3.77 & 0.61 & 2.94 & 4.98 & 9.29 & 1.96 & 3.37 & 6.38 & 9.65 \\
 & BERT-Attack & \cellcolor[HTML]{EFEFEF}1.10 & \cellcolor[HTML]{EFEFEF}11.04 & \cellcolor[HTML]{EFEFEF}11.59 & \cellcolor[HTML]{EFEFEF}24.86 & 0.90 & 10.22 & 10.10 & 21.56 & 6.20 & 19.95 & 27.65 & 41.18 & 8.55 & 22.16 & 30.02 & 42.44 \\
 & Sep-Attack & \cellcolor[HTML]{EFEFEF}44.60 & \cellcolor[HTML]{EFEFEF}55.78 & \cellcolor[HTML]{EFEFEF}64.45 & \cellcolor[HTML]{EFEFEF}71.65 & 2.40 & 12.69 & 13.88 & 25.54 & 6.61 & 19.84 & 28.07 & 41.14 & 8.65 & 21.87 & 30.24 & 42.59 \\
 & Co-Attack & \cellcolor[HTML]{EFEFEF}45.00 & \cellcolor[HTML]{EFEFEF}56.25 & \cellcolor[HTML]{EFEFEF}65.19 & \cellcolor[HTML]{EFEFEF}72.39 & 2.30 & 12.73 & 13.98 & 25.39 & 6.20 & 20.19 & 27.77 & 41.23 & 8.86 & 22.14 & 30.62 & 42.61 \\
 & SGA & \cellcolor[HTML]{EFEFEF}91.90 & \cellcolor[HTML]{EFEFEF}92.44 & \cellcolor[HTML]{EFEFEF}90.35 & \cellcolor[HTML]{EFEFEF}91.11 & 16.13 & 27.82 & 30.60 & 38.91 & 8.64 & 21.00 & 31.41 & 43.00 & 10.50 & 22.97 & 34.97 & 45.06 \\
 & DRA & \cellcolor[HTML]{EFEFEF}90.00 & \cellcolor[HTML]{EFEFEF}91.61 & \cellcolor[HTML]{EFEFEF}88.70 & \cellcolor[HTML]{EFEFEF}90.02 & 17.84 & 29.81 & 32.91 & 45.34 & 8.84 & 22.76 & 32.95 & 43.35 & 11.02 & 24.92 & 35.72 & 46.69 \\
 & SA-AET & \cellcolor[HTML]{EFEFEF}91.90 & \cellcolor[HTML]{EFEFEF}93.13 & \cellcolor[HTML]{EFEFEF}94.61 & \cellcolor[HTML]{EFEFEF}95.29 & 24.55 & 36.74 & 42.47 & 51.70 & 10.96 & 25.51 & 34.75 & 46.48 & 13.21 & 28.37 & 38.73 & 51.07 \\
\multirow{-8}{*}{ALBEF} & SimVLA & \cellcolor[HTML]{EFEFEF}83.10 & \cellcolor[HTML]{EFEFEF}86.41 & \cellcolor[HTML]{EFEFEF}86.95 & \cellcolor[HTML]{EFEFEF}87.74 & \textbf{31.46} & \textbf{45.60} & \textbf{49.86} & \textbf{58.75} & \textbf{17.58} & \textbf{33.32} & \textbf{46.55} & \textbf{55.92} & \textbf{21.65} & \textbf{39.69} & \textbf{51.03} & \textbf{60.72} \\ \hline
 & PGD & 1.30 & 2.85 & 3.31 & 5.84 & \cellcolor[HTML]{EFEFEF}64.43 & \cellcolor[HTML]{EFEFEF}66.35 & \cellcolor[HTML]{EFEFEF}65.70 & \cellcolor[HTML]{EFEFEF}67.72 & 0.30 & 3.07 & 5.02 & 9.43 & 1.54 & 3.32 & 6.36 & 9.51 \\
 & BERT-Attack & 0.70 & 10.80 & 11.65 & 24.46 & \cellcolor[HTML]{EFEFEF}0.60 & \cellcolor[HTML]{EFEFEF}10.87 & \cellcolor[HTML]{EFEFEF}12.29 & \cellcolor[HTML]{EFEFEF}23.85 & 7.72 & 21.02 & 28.83 & 41.86 & 9.37 & 22.61 & 31.86 & 43.28 \\
 & Sep-Attack & 3.80 & 15.89 & 17.41 & 31.41 & \cellcolor[HTML]{EFEFEF}67.23 & \cellcolor[HTML]{EFEFEF}73.31 & \cellcolor[HTML]{EFEFEF}76.60 & \cellcolor[HTML]{EFEFEF}80.47 & 8.54 & 20.58 & 29.12 & 41.62 & 9.47 & 22.88 & 32.03 & 43.40 \\
 & Co-Attack & 3.70 & 16.40 & 17.09 & 61.63 & \cellcolor[HTML]{EFEFEF}70.04 & \cellcolor[HTML]{EFEFEF}73.90 & \cellcolor[HTML]{EFEFEF}75.80 & \cellcolor[HTML]{EFEFEF}80.49 & 7.93 & 20.69 & 29.14 & 41.59 & 9.78 & 22.90 & 31.72 & 43.45 \\
 & SGA & 23.40 & 34.61 & 40.33 & 50.93 & \cellcolor[HTML]{EFEFEF}95.19 & \cellcolor[HTML]{EFEFEF}95.78 & \cellcolor[HTML]{EFEFEF}96.55 & \cellcolor[HTML]{EFEFEF}97.30 & 9.55 & 21.72 & 31.96 & 42.44 & 11.43 & 23.61 & 34.89 & 44.61 \\
 & DRA & 25.00 & 37.51 & 42.28 & 53.85 & \cellcolor[HTML]{EFEFEF}93.39 & \cellcolor[HTML]{EFEFEF}94.90 & \cellcolor[HTML]{EFEFEF}95.66 & \cellcolor[HTML]{EFEFEF}96.75 & 10.16 & 23.99 & 34.75 & 46.07 & 12.67 & 25.77 & 37.32 & 47.78 \\
 & SA-AET & 28.96 & 42.00 & 46.98 & 57.33 & \cellcolor[HTML]{EFEFEF}93.40 & \cellcolor[HTML]{EFEFEF}94.18 & \cellcolor[HTML]{EFEFEF}94.03 & \cellcolor[HTML]{EFEFEF}95.13 & 10.87 & 24.79 & 36.11 & 46.76 & 14.84 & 29.09 & 38.90 & 52.01 \\
\multirow{-8}{*}{TCL} & SimVLA & \textbf{38.38} & \textbf{51.44} & \textbf{59.14} & \textbf{67.27} & \cellcolor[HTML]{EFEFEF}90.60 & \cellcolor[HTML]{EFEFEF}91.97 & \cellcolor[HTML]{EFEFEF}92.94 & \cellcolor[HTML]{EFEFEF}94.38 & \textbf{18.70} & \textbf{34.93} & \textbf{48.66} & \textbf{56.48} & \textbf{22.26} & \textbf{40.06} & \textbf{52.07} & \textbf{61.70} \\ \hline
 & PGD & 0.10 & 0.97 & 1.75 & 3.05 & 0.00 & 1.24 & 2.09 & 3.48 & \cellcolor[HTML]{EFEFEF}41.77 & \cellcolor[HTML]{EFEFEF}49.24 & \cellcolor[HTML]{EFEFEF}66.22 & \cellcolor[HTML]{EFEFEF}69.56 & 0.72 & 1.33 & 3.79 & 5.73 \\
 & BERT-Attack & 0.40 & 8.07 & 8.17 & 20.27 & 0.60 & 8.14 & 8.73 & 19.86 & \cellcolor[HTML]{EFEFEF}6.71 & \cellcolor[HTML]{EFEFEF}17.42 & \cellcolor[HTML]{EFEFEF}29.00 & \cellcolor[HTML]{EFEFEF}38.74 & 5.77 & 18.68 & 28.95 & 39.41 \\
 & Sep-Attack & 0.60 & 7.97 & 8.21 & 20.61 & 0.60 & 8.85 & 9.08 & 20.50 & \cellcolor[HTML]{EFEFEF}52.03 & \cellcolor[HTML]{EFEFEF}63.74 & \cellcolor[HTML]{EFEFEF}78.20 & \cellcolor[HTML]{EFEFEF}82.77 & 7.62 & 20.94 & 30.02 & 40.34 \\
 & Co-Attack & 0.60 & 8.73 & 8.79 & 21.37 & 0.60 & 8.47 & 9.54 & 21.43 & \cellcolor[HTML]{EFEFEF}76.22 & \cellcolor[HTML]{EFEFEF}85.24 & \cellcolor[HTML]{EFEFEF}92.85 & \cellcolor[HTML]{EFEFEF}94.32 & 6.59 & 19.25 & 31.50 & 42.42 \\
 & SGA & 1.30 & 9.81 & 10.91 & 22.43 & 1.00 & 10.95 & 11.67 & 23.94 & \cellcolor[HTML]{EFEFEF}94.92 & \cellcolor[HTML]{EFEFEF}95.90 & \cellcolor[HTML]{EFEFEF}98.98 & \cellcolor[HTML]{EFEFEF}98.95 & 11.53 & 24.33 & 36.86 & 47.10 \\
 & DRA & 1.60 & 10.57 & 12.25 & 25.51 & 1.20 & 11.58 & 12.27 & 26.14 & \cellcolor[HTML]{EFEFEF}93.60 & \cellcolor[HTML]{EFEFEF}94.74 & \cellcolor[HTML]{EFEFEF}98.32 & \cellcolor[HTML]{EFEFEF}98.72 & 13.70 & 27.74 & 41.83 & 50.73 \\
 & SA-AET & 1.60 & 11.12 & 12.60 & 26.07 & 1.60 & 11.84 & 12.74 & 26.64 & \cellcolor[HTML]{EFEFEF}94.00 & \cellcolor[HTML]{EFEFEF}95.49 & \cellcolor[HTML]{EFEFEF}97.89 & \cellcolor[HTML]{EFEFEF}98.48 & 15.96 & 27.67 & 41.49 & 51.45 \\
\multirow{-8}{*}{$\rm CLIP_{ViT}$} & SimVLA & \textbf{4.20} & \textbf{20.06} & \textbf{19.41} & \textbf{36.97} & \textbf{4.91} & \textbf{21.67} & \textbf{20.29} & \textbf{38.69} & \cellcolor[HTML]{EFEFEF}82.32 & \cellcolor[HTML]{EFEFEF}87.22 & \cellcolor[HTML]{EFEFEF}94.53 & \cellcolor[HTML]{EFEFEF}95.96 & \textbf{22.97} & \textbf{38.22} & \textbf{54.96} & \textbf{64.07} \\ \hline
 & PGD & 0.10 & 0.79 & 1.56 & 2.77 & 0.00 & 1.11 & 1.86 & 3.41 & 0.41 & 1.42 & 1.75 & 4.69 & \cellcolor[HTML]{EFEFEF}62.31 & \cellcolor[HTML]{EFEFEF}73.98 & \cellcolor[HTML]{EFEFEF}76.00 & \cellcolor[HTML]{EFEFEF}84.25 \\
 & BERT-Attack & 0.60 & 8.29 & 7.95 & 20.52 & 0.70 & 8.29 & 8.30 & 20.12 & 6.71 & 17.64 & 27.46 & 37.00 & \cellcolor[HTML]{EFEFEF}7.11 & \cellcolor[HTML]{EFEFEF}20.89 & \cellcolor[HTML]{EFEFEF}30.91 & \cellcolor[HTML]{EFEFEF}41.83 \\
 & Sep-Attack & 0.70 & 8.29 & 8.07 & 20.46 & 0.60 & 8.55 & 8.48 & 20.67 & 6.81 & 18.47 & 27.65 & 38.50 & \cellcolor[HTML]{EFEFEF}71.16 & \cellcolor[HTML]{EFEFEF}81.57 & \cellcolor[HTML]{EFEFEF}86.48 & \cellcolor[HTML]{EFEFEF}91.55 \\
 & Co-Attack & 0.60 & 8.45 & 8.05 & 20.60 & 0.80 & 8.71 & 8.73 & 20.81 & 6.50 & 18.86 & 28.17 & 38.60 & \cellcolor[HTML]{EFEFEF}80.02 & \cellcolor[HTML]{EFEFEF}87.35 & \cellcolor[HTML]{EFEFEF}90.78 & \cellcolor[HTML]{EFEFEF}94.76 \\
 & SGA & 1.10 & 9.14 & 9.39 & 21.50 & 1.20 & 9.93 & 10.63 & 22.60 & 8.43 & 20.17 & 32.72 & 40.52 & \cellcolor[HTML]{EFEFEF}95.88 & \cellcolor[HTML]{EFEFEF}97.31 & \cellcolor[HTML]{EFEFEF}98.42 & \cellcolor[HTML]{EFEFEF}98.82 \\
 & DRA & 1.20 & 9.75 & 9.88 & 23.44 & 1.10 & 10.64 & 10.36 & 24.08 & 9.25 & 22.22 & 35.16 & 43.49 & \cellcolor[HTML]{EFEFEF}94.85 & \cellcolor[HTML]{EFEFEF}96.43 & \cellcolor[HTML]{EFEFEF}97.82 & \cellcolor[HTML]{EFEFEF}98.26 \\
 & SA-AET & 1.30 & 9.95 & 10.19 & 23.51 & 1.30 & 10.89 & 11.03 & 24.32 & 10.09 & 22.93 & 36.76 & 45.83 & \cellcolor[HTML]{EFEFEF}97.32 & \cellcolor[HTML]{EFEFEF}97.54 & \cellcolor[HTML]{EFEFEF}98.48 & \cellcolor[HTML]{EFEFEF}99.05 \\
\multirow{-8}{*}{$\rm CLIP_{CNN}$} & SimVLA & \textbf{3.00} & \textbf{18.92} & \textbf{18.28} & \textbf{35.11} & \textbf{3.91} & \textbf{21.41} & \textbf{18.80} & \textbf{37.81} & \textbf{16.99} & \textbf{33.36} & \textbf{51.01} & \textbf{58.19} & \cellcolor[HTML]{EFEFEF}91.35 & \cellcolor[HTML]{EFEFEF}94.44 & \cellcolor[HTML]{EFEFEF}97.02 & \cellcolor[HTML]{EFEFEF}97.80 \\ \toprule[1pt]
\end{tabular}
}
\end{table*}

\begin{table}[!htb] 
	\caption{Evaluation of transferability on BLIP (using ViT-B as the image encoder), with Flickr30k as the dataset.}
	\label{blip}
	\centering
	\small
	\setlength{\tabcolsep}{2.5mm}
\begin{tabular}{cccccc}
\toprule[1pt]
 &  & \multicolumn{2}{c}{TR} & \multicolumn{2}{c}{IR} \\
\multirow{-2}{*}{Model} & \multirow{-2}{*}{Attack} & R@1 & R@10 & R@1 & R@10 \\ \toprule[1pt]
 & SGA & 36.07 & 12.20 & 45.59 & 22.94 \\
 & DRA & 37.18 & 14.02 & 49.39 & 25.40 \\
 & {SA-AET} & 43.31 & 16.97 & 54.48 & 29.87 \\
\multirow{-4}{*}{ALBEF} & SimVLA & \textbf{57.18} & \textbf{27.34} & \textbf{65.27} & \textbf{42.87} \\ \hline
 & SGA & 13.62 & 2.03 & 27.19 & 11.38 \\
 & DRA & 15.21 & 2.44 & 29.25 & 12.82 \\
 & {SA-AET} & 14.23 & 2.34 & 29.54 & 12.84 \\
\multirow{-4}{*}{$\rm CLIP_{CNN}$} & SimVLA & \textbf{25.52} & \textbf{6.00} & \textbf{44.43} & \textbf{23.66} \\ \toprule[1pt]
	\end{tabular}
	
\end{table}

\begin{table*}[!htb] 
	\caption{
	Transferability comparison under image perturbation bounds $\epsilon_{x}=8/255$ and $\epsilon_{x}=16/255$, with ALBEF ($\rm CLIP_{ViT}$) as the surrogate (target) model on Flickr30k.}
\label{bound_compare}
\centering
	 \small 
	\setlength{\tabcolsep}{6.0mm} 
\begin{tabular}{ccccccccc}
\toprule[1pt]
 & \multicolumn{4}{c}{$\epsilon_{x}=8/255$} & \multicolumn{4}{c}{$\epsilon_{x}=16/255$} \\ \cmidrule(lr){2-5} \cmidrule(lr){6-9}
 & \multicolumn{2}{c}{TR} & \multicolumn{2}{c}{IR} & \multicolumn{2}{c}{TR} & \multicolumn{2}{c}{IR} \\
\multirow{-3}{*}{Attack} &  R@1 & R@10 & R@1 & R@10 & R@1 & R@10 & R@1 & R@10 \\ \toprule[1pt]
{SGA} & 42.94 & 15.96 & 55.19 & 28.65 & 46.99 & 19.82 & 57.86 & 31.01 \\
{DRA} & 44.79 & 16.97 & 57.49 & 29.83 & 57.06 & 27.44 & 65.56 & 38.73 \\
{SA-AET} & 55.21 & 26.22 & 63.56 & 36.68 & 71.25 & 40.99 & 76.61 & 52.53 \\
{SimVLA} & \textbf{63.31} & \textbf{29.78} & \textbf{70.04} & \textbf{44.13} & \textbf{72.15} & \textbf{41.67} & \textbf{77.16} & \textbf{54.21} \\ \toprule[1pt]
\end{tabular}
\end{table*}

\begin{table}[t] 
	\caption{
	Cross-model-and-task transferability (\%) from Retrieval to Visual Entailment (VE) on a 2K subset of SNLI-VE.}
\label{ve}
	\centering
    \small
	\setlength{\tabcolsep}{4.0mm}
\begin{tabular}{cccc}
\toprule[1pt]
Model & Attack & ALBEF (VE) & TCL (VE) \\ \toprule[1pt]
\multirow{4}{*}{$\rm CLIP_{ViT}$} & SGA & 32.97 & 33.32 \\
 & DRA & 32.70 & 33.77 \\
 & {SA-AET} & 33.17 & 34.02 \\
 & SimVLA & \textbf{38.97} & \textbf{38.67} \\ \hline
\multirow{4}{*}{$\rm CLIP_{CNN}$} & SGA & 32.72 & 33.42 \\
 & DRA & 32.52 & 33.72 \\
 & {SA-AET} & 32.37 & 33.67 \\
 & SimVLA & \textbf{39.07} & \textbf{38.12} \\ \toprule[1pt]
\end{tabular}
\end{table}

\begin{table}[t] 
	\caption{
	Cross-model-and-task transferability (\%) from Retrieval to Visual Grounding (VG) on three splits of RefCOCO+ dataset: Val, TestA, and TestB.}
\label{vg}
	\centering
    \small
	\setlength{\tabcolsep}{2.0mm}
\begin{tabular}{ccccc}
\toprule[1pt]
Model & Attack & Val (VG) & TestA (VG) & TestB (VG) \\ \toprule[1pt]
\multirow{4}{*}{$\rm CLIP_{ViT}$} & SGA & 54.92 & 50.00 & 61.08 \\
 & DRA & 55.12 & 49.56 & 61.01 \\
 & SA-AET & 55.04 & 49.63 & \textbf{61.61} \\
 & SimVLA & \textbf{56.93} & \textbf{52.83} & \textbf{61.61} \\ \hline
\multirow{4}{*}{$\rm CLIP_{CNN}$} & SGA & 54.64 & 49.69 & 61.24 \\
 & DRA & 54.70 & 49.62 & 61.35 \\
 & SA-AET & 54.68 & 49.51 & \textbf{61.38} \\
 & SimVLA & \textbf{55.85} & \textbf{51.10} & 61.12 \\ \toprule[1pt]
\end{tabular}
\end{table}

\begin{table*}[!t] 
	\caption{Attack performance under image and text defenses, with ALBEF ($\rm CLIP_{ViT}$) as the surrogate (target) models and Flickr30k as the dataset.}
	\label{defense}
\centering
\small
\setlength{\tabcolsep}{1.3mm}
\begin{tabular}{ccccccccc}
\toprule[1pt]
 & \multicolumn{4}{c}{JPEG (Image Defense)} & \multicolumn{4}{c}{Bit (Image Defense)} \\ \cmidrule(lr){2-5} \cmidrule(lr){6-9}
 & \multicolumn{2}{c}{TR} & \multicolumn{2}{c}{IR} & \multicolumn{2}{c}{TR} & \multicolumn{2}{c}{IR} \\
\multirow{-3}{*}{Attack} & R@1 & R@10 & R@1 & R@10 &  R@1 & R@10 & R@1 & R@10 \\ \toprule[1pt]
SGA & 32.64 (+0.25) & 10.26 (+1.62) & 45.91 (+1.84) & 21.19 (+0.19) & 33.01 (+0.62) & 9.04 (+0.40) & 45.78 (+1.71) & 21.81 (+0.81) \\
DRA & 34.60 (-2.09) & 10.06 (+1.22) & 49.10 (+1.13) & 23.44 (+0.68) & 34.97 (-1.72) & 9.25 (+0.41) & 48.90 (+0.93) & 23.35 (+0.59) \\
SA-AET & 37.06 (-0.75) & 10.77 (-0.19) & 49.61 (-0.87) & 23.90 (-1.61) & 35.71 (-2.10) & 10.26 (-0.70) & 50.00 (-0.48) & 24.05 (-1.46) \\
SimVLA & \textbf{49.69 (-2.83)} & \textbf{16.87 (-0.71)} & \textbf{61.50 (+0.1.26)} & \textbf{34.06 (+0.74)} & \textbf{50.67 (-1.85)} & \textbf{16.87 (-0.71)} & \textbf{60.50 (+0.26)} & \textbf{33.62 (+0.30)} \\ \toprule[1pt]
 & \multicolumn{4}{c}{Substitution (Text Defense)} & \multicolumn{4}{c}{Substitution (Text Defense)} \\ \cmidrule(lr){2-5} \cmidrule(lr){6-9}
 & \multicolumn{2}{c}{TR} & \multicolumn{2}{c}{IR} & \multicolumn{2}{c}{TR} & \multicolumn{2}{c}{IR} \\
\multirow{-3}{*}{Attack} &  R@1 & R@10 & R@1 & R@10 &  R@1 & R@10 & R@1 & R@10 \\ \toprule[1pt]
SGA & 30.31 (-2.08) & 7.42 (-1.22) & 42.85 (-1.22) & 20.48 (-0.52) & 27.85 (-4.54) & 6.50 (-2.14) & 39.66 (-4.41) & 17.77 (-3.23) \\
DRA & 34.60 (-2.09) & 7.93 (-0.91) & 45.91 (-2.06) & 22.22 (-0.54) & 30.80  (-5.89) & 6.50 (-2.34) & 42.59 (-5.38) & 19.30 (-3.46) \\
SA-AET & 35.21 (-2.60) & 8.74 (-2.22) & 47.62 (-2.86) & 22.29 (-3.22) & 32.39 (-5.42) & 7.52 (-3.44) & 44.43 (-6.05) & 20.24 (-5.27) \\
SimVLA & \textbf{42.09 (-10.43)} & \textbf{13.11 (-4.47)} & \textbf{54.83 (-5.41)} & \textbf{27.95 (-5.37)} & \textbf{39.75 (-12.77)} & \textbf{10.77 (-6.81)} & \textbf{51.90 (-8.34)} & \textbf{25.99 (-7.33)} \\ \toprule[1pt]
\end{tabular}
\end{table*}

\begin{table*}[!t] 
	\caption{Attack performance on robust VLPMs, with ALBEF ($\rm CLIP_{ViT}$) as the surrogate (target) models and Flickr30k as the dataset.}
	\label{robust_vlp}
\centering
\small
\setlength{\tabcolsep}{6mm}
\begin{tabular}{ccccccccc}
\toprule[1pt]
\multirow{3}{*}{Attack} & \multicolumn{4}{c}{TeCoA} & \multicolumn{4}{c}{FARE} \\ \cmidrule(lr){2-5} \cmidrule(lr){6-9}
 & \multicolumn{2}{c}{TR} & \multicolumn{2}{c}{IR} & \multicolumn{2}{c}{TR} & \multicolumn{2}{c}{IR} \\
 & R@1 & R@10 & R@1 & R@10 & R@1 & R@10 & R@1 & R@10 \\ \toprule[1pt]
SGA & 58.00 & 22.50 & 70.60 & 35.24 & 41.60 & 9.10 & 61.76 & 25.42 \\
DRA & 53.70 & 20.40 & 66.46 & 31.42 & 38.20 & 8.70 & 58.34 & 22.60 \\
SA-AET & 53.10 & 20.40 & 66.52 & 31.28 & 38.60 & 9.30 & 59.02 & 22.76 \\
SimVLA & \textbf{64.60} & \textbf{28.20} & \textbf{75.32} & \textbf{41.44} & \textbf{52.20} & \textbf{15.20} & \textbf{68.36} & \textbf{31.98} \\ \toprule[1pt]
\end{tabular}
\end{table*}

\section{Experiments}
In this section, we compare our new attack pipeline, SimVLA, to six attack baselines (that have publicly released their codes) across three tasks on four datasets and with four VLPMs. We evaluate, the (cross-model and cross-task) transferability, computational efficiency and attack stealthiness.
Finally, we conduct ablation studies of the main components of SimVLA and parameter analysis.
\subsection{Experimental Settings}
\subsubsection{Tasks and Datasets} 
Following the common practice \cite{coattack,sga,dra}, we mainly focus on Text Retrieval (TR) and Image Retrieval (IR) tasks and also consider visual entailment (VE) and visual grounding (VG).
Both TR and IR are conducted on a 1K subset of the Flickr30k (1k images paired with 5k captions) \cite{fk} and a 5K subset of MSCOCO (5k images paired with 25k captions) \cite{coco} datasets.
VE is conducted on a 2K subset of SNLI-VE dataset \cite{ve}, and VG is conducted on the validation and test sets of RefCOCO+ dataset \cite{vg}, involving about 21k text-image pairs.

\subsubsection{Evaluation Metrics}
Transferability refers to the attack success rates (\%) on black-box target models.
For TR and IR tasks, R@k is calculated as the attack success rate, which measures the proportion of originally correct matches that no longer rank within the top k results after the attack.
In particular, we follow the common practice of reporting the R@1 and R@10.
For the VE task, the attack is considered successful only if the entailment relationship of an adversarial pair changes.
For the VG task, the attack is considered successful only if the IoU of the grounding regions between the adversarial and clean pairs is lower than 0.5.
In addition to transferability, we also evaluate the efficiency regarding the computational time and VRAM. 
For evaluation on MLLM, evaluation details are provided in Appendix \ref{mllm_eval_details}. 

\subsubsection{Models} 
We follow previous works (Co-Attack \cite{coattack}, SGA \cite{sga}, and DRA \cite{dra}), mainly focusing on four widely used VLPMs as both surrogate and target models: two fused VLPMs, ALBEF \cite{albef} and TCL \cite{tcl}, and two aligned VLPMs, $\rm CLIP_{ViT}$ and $\rm CLIP_{CNN}$. All four models can handle IR and TR tasks.
TCL can handle the VE task, while ALBEF can handle both VE and VG tasks.
In addition, we further evaluate our attack pipeline on the popular VLPM BLIP \cite{blip}; adversarially robust VLPMs, including TeCoA \cite{tecoa} and FARE \cite{fare}; popular open-source multimodal large language models (MLLMs), including Qwen2.5-VL-3B \cite{qwenvl}, Qwen2.5-VL-7B \cite{qwenvl}, and LLaVA-Mistral-7B \cite{llava}; and closed-source MLLMs, including GPT-4.1-Nano \cite{chatgpt}, GPT-4.1-Mini \cite{chatgpt}, GPT-5-Nano \cite{chatgpt}, GPT-5-Mini \cite{chatgpt}, GPT-5.4 \cite{chatgpt}, and GPT-5.5 \cite{chatgpt}.

\subsubsection{Baselines} 
Following the latest work~\cite{dra}, we adopt previously published and well-established open-source attacks and pipelines as our baselines to ensure the reproducibility and accuracy of our results, including PGD (ICLR 2018) \cite{pgd}, BERT-Attack (EMNLP 2020) \cite{BERTattack}, Sep-Attack (ACM MM 2022) \cite{coattack}, Co-Attack (ACM MM 2022) \cite{coattack}, SGA (ICCV 2023) \cite{sga}, DRA (ECCV 2024) \cite{dra}, and current SOTA SA-AET (TPAMI 2025) \cite{saaet}.

\subsubsection{Parameter Settings} 
We follow the common practice when setting the main hyperparameters \cite{coattack,sga,sa,ot,contrastattack}.
Specifically, for image attacks, the total number of iterations is $T=10$, the image perturbation bound is $\epsilon_{x}=2/255$ and the step size is $\alpha=\epsilon_{x}/4=0.5/255$.
For text attacks, previous works did not restrict the semantics of the replacement words, but only limited the number of replacements. Therefore, the commonly choice of the maximum number of replaced words is $\epsilon_{t}=1$.
For the transferable attack baselines, we adopt their default settings.
For our SimVLA, the boundary factor $\beta$ for the sampled image in the cross-modal word identification of Stage 1 is 16, the number of removed words in the text semantic abstraction of Stage 2 is Top-$K=1$,
and the sample number is $N_I=4$, consistent with the number of augmented images in SGA \cite{sga} and DRA \cite{dra}. 
All experiments for SimVLA and other baseline methods are conducted on an RTX 3090 GPU (24 GB), except for the reproduction of the SA-AET baseline, which is performed on an L40s GPU (48 GB). The random seed was set to 42, and the batch size was fixed to 2. The code will be released as soon as possible upon the final decision.

\subsection{Experimental Results}
\subsubsection{Cross-model Transferability} 
Following previous work \cite{coattack,sga,dra}, we mainly evaluate the transferability in retrieval tasks, consisting of text retrieval (TR) and image retrieval (IR).
As reported in \autoref{body_r1} and \autoref{body_r10}, our attack pipeline achieves the highest transferability on both Flickr30k and MSCOCO datasets regarding both R@1 and R@10 metrics. For instance, on Flickr30k for TR, the R@10 transferability of SimVLA reaches 17.58\%, almost doubling the performance of the previous attack pipeline, DRA. 
Note that the white-box success is not the focus of transferable attacks~\cite{inkawhich2019feature,logit,springer2021little}, including ours.

Besides, we also provide additional results on a mainstream VLPM BLIP \cite{blip}. 
The results in \autoref{blip} demonstrate that our SimVLA consistently outperforms SGA, DRA, and SA-AET in both Text Retrieval (TR) and Image Retrieval (IR) tasks.

Finally, we provide the transferability comparison of advanced baselines under larger image perturbation bound, i.e., $\epsilon_{x}=8/255$ and $\epsilon_{x}=16/255$. As shown in \autoref{bound_compare}, our SimVLA is still the best.

\subsubsection{Cross-Model-and-Task Transferability} 
We follow recent works \cite{sga,sa,ot} to consider transferable attacks on both cross-model and cross-task settings.
Specifically, we generate adversarial examples on a surrogate model for the retrieval task and directly apply these pairs to attacking unknown target models for visual grounding (VG) and visual entailment (VE) tasks.
As shown in \autoref{ve} and \autoref{vg}, SimVLA still achieves the best results in nearly all cases. 
For instance, transferring from $\rm CLIP_{ViT}$ (Retrieval) to ALBEF (VE), yields an attack success rate of 39.07\%, surpassing the previous best attack pipeline, DRA, by 6.55\%.

\subsubsection{Attack under Defenses}  
We evaluate different attacks under four widely used defense strategies. Specifically, JPEG \cite{jpeg} and Bit \cite{bit} mitigate adversarial perturbations by distorting the input, and are commonly adopted in traditional transfer-based attacks \cite{gra,vt,fpr,mumodig,svre,ssa}. Substitution and Deletion assume that the defender has knowledge of the adversarial word positions and either replaces them with random strings or removes them entirely, thereby precisely eliminating the adverse effects of adversarial words.

Results from \autoref{defense} show that SimVLA consistently achieves the best results. Additionally, we observe that image defenses targeting traditional image adversarial perturbation, such as JPEG and Bit, perform poorly, whereas text defenses reduce attack ASR the most in multimodal retrieval. This suggests that VLPMs rely heavily on cross-modal semantic alignment between text and image representations, making low-level pixel distortions less effective, while disrupting textual semantics directly undermines the learned text--image alignment that underpins retrieval performance

We also evaluate the performance of our SimVLA on two adversarially robust VLPMs: TeCoA \cite{tecoa} and FARE \cite{fare}. Results in \autoref{robust_vlp} show that SimVLA is still the best.

\subsubsection{Attack on MLLMs}  
Beyond VLPMs, we evaluate various attacks on popular open-source MLLMs, including Qwen2.5-VL-3B \cite{qwenvl}, Qwen2.5-VL-7B \cite{qwenvl} and LLaVA-Mistral-7B \cite{llava}. In this process, we send the adversarial text-image pairs (5k text-image pairs per attack) to the target black-box MLLM and ask whether the input pair matches in terms of content.
\autoref{mllm_local} shows that our attack pipeline yields the highest attack success rate.

Moreover, we further evaluate the performance of SimVLA on closed-source  MLLMs. We first evaluate on four cost-effective lightweight APIs, including GPT4.1-Nano \cite{chatgpt}, GPT4.1-Mini \cite{chatgpt}, GPT5-Nano \cite{chatgpt}, and GPT5-Mini \cite{chatgpt}.
The results in \autoref{mllm_light_api} indicate that models from the GPT5 series exhibit stronger robustness than those from GPT4.1, and SimVLA provides a more effective means of assessing model robustness compared to other advanced attacks.

We further consider two advanced closed-source APIs, GPT5.4 \cite{chatgpt} and GPT5.5 \cite{chatgpt}. Due to the high evaluation cost, we select the first 1K adversarial text-image pairs generated from the dataset. As shown in Table XII, our method substantially outperforms the baselines.

\subsubsection{Computational Efficiency}
In addition to the above evaluation of transferability, we also evaluate the computational efficiency of different attack pipelines, since the practical considerations are important for black-box attacks.
As can be seen from \autoref{cost}, our SimVLA achieves the highest efficiency since it gets rid of the unnecessary Stage 3 and adopt a simple loss function.
Notably, SimVLA requires only 50\% of the computation time and 60\% of the max VRAM compared to the previous state-of-the-art pipeline, DRA.
This further indicates that genuinely effective modal interaction strategies are always simple rather than multiple complex interaction stages and complicated loss functions.

\begin{table}[t] 
	\caption{
	Attack performance on three open-source MLLMs, with ALBEF as the surrogate  models and Flickr30k as the dataset.}
    \label{mllm_local}
    \centering
    \small
    	\setlength{\tabcolsep}{0.9mm}
\begin{tabular}{cccc}
\toprule[1pt]
Attack & Qwen2.5-VL-3B & Qwen2.5-VL-7B & LLaVA-Mistral-7B \\ \toprule[1pt]
SGA & 64.70 & 66.60 & 91.08 \\
DRA & 50.34 & 51.12 & 90.14 \\
SA-AET & 52.28 & 51.38 & 90.20 \\
SimVLA & \textbf{68.80} & \textbf{70.58} & \textbf{94.12} \\ \toprule[1pt]
\end{tabular}
\end{table}

\begin{table}[t] 
	\caption{
	Attack performance on the APIs of four lightweight closed-source  MLLMs, with ALBEF as the surrogate  models and Flickr30k as the dataset.}
    \label{mllm_light_api}
    \centering
    \small
    	\setlength{\tabcolsep}{0.9mm}
\begin{tabular}{ccccc}
\toprule[1pt]
\multicolumn{1}{l}{Attack} & \multicolumn{1}{l}{GPT4.1-Nano} & \multicolumn{1}{l}{GPT4.1-Mini} & \multicolumn{1}{l}{GPT5-Nano} & \multicolumn{1}{l}{GPT5-Mini} \\ \toprule[1pt]
SGA & 69.01 & 68.57 & 32.92 & 47.44 \\
DRA & 61.48 & 61.58 & 18.06 & 22.86 \\
SA-AET & 62.03 & 61.44 & 17.96 & 22.66 \\
SimVLA & \textbf{72.76} & \textbf{72.43} & \textbf{34.57} & \textbf{49.53} \\ \toprule[1pt]
\end{tabular}
\end{table}

\begin{table}[t] 
	\caption{
	Attack performance on the APIs of two advanced closed-source  MLLMs, with ALBEF as the surrogate model and Flickr30k (the first 1k part) as the dataset.}
    \label{mllm_advanced_api}
    \centering
    \small
    \setlength{\tabcolsep}{9.2mm}
\begin{tabular}{ccc}
\toprule[1pt]
\multicolumn{1}{l}{Attack} &
\multicolumn{1}{l}{GPT5.4} &
\multicolumn{1}{l}{GPT5.5} \\ 
\toprule[1pt]
SGA    & 49.70 & 41.03 \\
DRA    & 32.20 & 19.57 \\
SA-AET & 31.50 & 19.24 \\
SimVLA & \textbf{56.40} & \textbf{45.16} \\
\toprule[1pt]
\end{tabular}
\end{table}

\begin{table}[t] 
	\caption{
		Computational efficiency.
		The batch size is 1, with ALBEF ($\rm CLIP_{ViT}$) as the surrogate (target) model on Flickr30k.}
	\label{cost}
\centering
\small
\setlength{\tabcolsep}{2.8mm}
\begin{tabular}{ccc}
\toprule[1pt]
Attack & Computation Time (s) $\downarrow$ & Max VRAM (MiB) $\downarrow$ \\ \toprule[1pt]
SGA & 3213 & 7180 \\
DRA & 6238 & 7292 \\
SA-AET & 8040 & 9737 \\
SimVLA & \textbf{2873} & \textbf{4504} \\ \toprule[1pt]
\end{tabular}
\end{table}

\begin{table}[!htb]
	\caption{Attack stealthiness. All approaches yield similar results.
	The dataset is Flickr30k, and the surrogate model is ALBEF.}
	\label{steal}
\centering
\small
\setlength{\tabcolsep}{7.mm}
\begin{tabular}{ccc}
\toprule[1pt]
Attack & Cosine Similarity $\uparrow$ & SSIM $\uparrow$ \\ \toprule[1pt]
SGA & 0.942 & \textbf{0.988} \\
DRA & 0.943 & \textbf{0.988} \\
SA-AET & 0.943 & 0.987 \\
SimVLA & \textbf{0.951} & 0.982 \\ \toprule[1pt]
\end{tabular}
\end{table}

\subsubsection{Attack Stealthiness}
All attack pipelines have been constrained by $\epsilon_{t}=1$ for texts and $\epsilon_{x}=2/255$ for images.
Here we compare their actual stealthiness. 
For the text, we calculate the semantic similarity between each original target word and its adversarial replacement.
For the image, we calculate the SSIM \cite{ssim} between each original image and its adversarial version.
\autoref{steal} shows that all three approaches achieve similar stealthiness.

\subsection{Ablation Studies}
We analyze the impact of components and parameters of our SimVLA on the transferability.
In ablation studies, the dataset adopted in all experiments is Flickr30k, surrogate model is ALBEF, target models are $\rm CLIP_{ViT}$, $\rm CLIP_{CNN}$ and TCL.



\subsubsection{Attack Components}
Our SimVLA refines the common attack pipeline in three aspects. As illustrated in \autoref{abl}, the cross-modal word identification (Stage 1) and stage deletion (Stage 3) contribute the most to the improved transferability, confirming that enriching cross-modal interactions and relieving the reliance on the surrogate model are important.
Text semantic abstraction (Stage 2) is also necessary but contributes less.
Note that when removing the refinement in Stage 2, SimVLA still substantially outperforms other baselines in \autoref{body_r1} and \autoref{body_r10}.
Furthermore, we find that MIM-PGIA is slightly better than PGD, and even incorporating MIM-PGIA, existing methods, SGA and DRA, still perform much worse than ours.
To disentangle the contributions of the three proposed components from that of the auxiliary MIM-PGIA optimizer, we further conduct controlled ablation experiments for each component under a setting without MIM-PGIA.  
As shown in \autoref{abl}, the results indicate that the three stage-wise refinements contribute to the overall performance gains, while MIM-PGIA provides an additional complementary benefit. Moreover, SimVLA without MIM-PGIA still consistently outperforms the current SOTA method, SA-AET-MIM-PGIA, across all twelve metrics.


\subsubsection{Order of Attack Stages} 
Our refined attack pipeline has removed the third stage.
We hypothesize that the order of the remaining two stages may also matter. 
To validate this, we test whether swapping their order, i.e., implementing the image attack (Stage 2) first would affect the attack performance.
In this case, we first generate adversarial images based on the clean text, and then generate adversarial text based on the adversarial images.
The results in \autoref{order} validate our hypothesis by showing that Text $\rightarrow$ Image is better.
This is because text attacks improve transferability more than image attacks (see BERT-Attack vs. PGD in \autoref{body_r1}).
Placing the text attack first immediately shifts the original text-image matching semantics.
Then, with the text as an effective target, the adversarial image moves in a more dissimilar direction, leading to a greater overall difference in the final adversarial text-image pair.

\begin{table*}[!t] 
	\caption{Ablation results of the main attack components, using ALBEF as the surrogate model. When each component is removed, the performance remains superior to others in most cases.}
	\label{abl}
\centering
\setlength{\tabcolsep}{1.4mm}
\small
\begin{tabular}{ccccccccccccc}
\toprule[1pt]
\multirow{3}{*}{Stragegy} & \multicolumn{4}{c}{$\rm CLIP_{ViT}$} & \multicolumn{4}{c}{ $\rm CLIP_{CNN}$} & \multicolumn{4}{c}{TCL} \\ \cmidrule(lr){2-5} \cmidrule(lr){6-9} \cmidrule(lr){10-13}
 & \multicolumn{2}{c}{TR} & \multicolumn{2}{c}{IR} & \multicolumn{2}{c}{TR} & \multicolumn{2}{c}{IR} & \multicolumn{2}{c}{TR} & \multicolumn{2}{c}{IR} \\
 & R@1 & R@10 & R@1 & R@10 & R@1 & R@10 & R@1 & R@10 & R@1 & R@10 & R@1 & R@10 \\ \toprule[1pt]
w/o refining Stage 1 & 41.60 & 13.62 & 56.60 & 29.68 & 50.18 & 14.43 & 61.69 & 34.45 & 60.85 & 28.66 & 70.12 & 42.59 \\
w/o MIM-PGIA \& w/o refining Stage 1 & 41.23 & 12.60 & 56.28 & 28.94 & 49.57 & 14.23 & 60.53 & 33.80 & 55.74 & 21.34 & 66.83 & 37.73 \\
w/o refining Stage 2 & 51.49 & 17.09 & 59.83 & 33.25 & \textbf{57.18} & 21.55 & \textbf{66.17} & 39.42 & 63.54 & 31.33 & 72.51 & \textbf{45.78} \\
w/o MIM-PGIA \& w/o refining Stage 2 & 50.55 & 16.97 & 58.70 & 32.93 & 55.83 & 20.73 & 65.50 & 38.75 & 58.48 & 25.55 & 68.90 & 40.98 \\
w/o refining Stage 3 & 45.77 & 16.16 & 56.64 & 30.44 & 51.78 & 19.41 & 62.69 & 36.09 & 62.65 & 30.46 & \textbf{72.90} & 45.02 \\
w/o MIM-PGIA \& w/o refining Stage 3 & 45.28 & 15.14 & 55.86 & 29.92 & 51.17 & 19.00 & 62.15 & 35.39 & 59.33 & 25.85 & 68.50 & 39.93 \\
w/o MIM-PGIA & 50.67 & 16.07 & 59.16 & 32.27 & 55.09 & 20.53 & 65.08 & 39.03 & 59.43 & 26.95 & 69.52 & 42.02 \\
SimVLA & \textbf{52.52} & \textbf{17.58} & \textbf{60.24} & \textbf{33.32} & 56.69 & \textbf{21.65} & 65.88 & \textbf{39.69} & \textbf{63.65} & \textbf{31.46} & 72.88 & 45.60 \\ \hline
SGA & 32.39 & 8.64 & 44.07 & 21.00 & 35.12 & 10.50 & 46.79 & 22.97 & 45.84 & 16.13 & 55.43 & 27.82 \\
SGA-MIM-PGIA & 34.11 & 9.76 & 45.75 & 21.94 & 37.11 & 11.36 & 46.96 & 23.36 & 47.79 & 18.05 & 57.54 & 29.65 \\
DRA & 36.69 & 8.84 & 47.79 & 22.76 & 39.59 & 11.02 & 49.78 & 24.92 & 48.16 & 17.84 & 57.40 & 29.81 \\
DRA-MIM-PGIA & 35.45 & 9.76 & 47.87 & 23.83 & 40.18 & 12.74 & 50.99 & 25.03 & 50.16 & 19.47 & 59.43 & 32.21 \\
SA-AET & 37.81 & 10.96 & 50.48 & 25.51 & 45.40 & 13.21 & 54.57 & 28.37 & 54.69 & 24.55 & 64.19 & 36.74 \\
SA-AET-MIM-PGIA & 38.16 & 11.57 & 49.84 & 25.53 & 46.50 & 13.82 & 55.67 & 28.96 & 55.95 & 25.65 & 65.05 & 37.23 \\ \toprule[1pt]
\end{tabular}
\end{table*}

\begin{table*}[!t] 
	\caption{Impact of the order of the first two attack stages, using ALBEF as the surrogate model, where Image $\rightarrow$ Text denotes the default order.}
	\label{order}
\centering
\small
\setlength{\tabcolsep}{2.6mm}
\begin{tabular}{ccccccccccccc}
\toprule[1pt]
\multirow{3}{*}{Order} & \multicolumn{4}{c}{$\rm CLIP_{ViT}$} & \multicolumn{4}{c}{$\rm CLIP_{CNN}$} & \multicolumn{4}{c}{TCL} \\ \cmidrule(lr){2-5} \cmidrule(lr){6-9} \cmidrule(lr){10-13}
 & \multicolumn{2}{c}{TR} & \multicolumn{2}{c}{IR} & \multicolumn{2}{c}{TR} & \multicolumn{2}{c}{IR} & \multicolumn{2}{c}{TR} & \multicolumn{2}{c}{IR} \\
 & R@1 & R@10 & R@1 & R@10 & R@1 & R@10 & R@1 & R@10 & R@1 & R@10 & R@1 & R@10 \\ \toprule[1pt]
Image $\xrightarrow{}$ Text & 41.60 & 12.91 & 51.10 & 33.64 & 49.94 & 16.77 & 58.34 & 32.25 & 62.44 & 30.57 & 71.88 & 43.74 \\
Text $\xrightarrow{}$ Image & \textbf{52.52} & \textbf{17.58} & \textbf{60.24} & \textbf{42.86} & \textbf{56.69} & \textbf{21.65} & \textbf{65.88} & \textbf{39.69} & \textbf{63.65} & \textbf{31.46} & \textbf{72.88} & \textbf{45.60} \\ \toprule[1pt]
\end{tabular}
\end{table*}

\begin{table*}[!t]
	\caption{The impact of the maximum number of replaced words $\epsilon_{t}$ on the attack success rate, using ALBEF as the surrogate model.}
	\label{txt_bound}
	\centering
	\small
	\setlength{\tabcolsep}{3.2mm}
\begin{tabular}{ccccccccccccc}
\toprule[1pt]
\multirow{3}{*}{$\epsilon_{t}$} & \multicolumn{4}{c}{$\rm CLIP_{ViT}$} & \multicolumn{4}{c}{$\rm CLIP_{CNN}$} & \multicolumn{4}{c}{TCL} \\ \cmidrule(lr){2-5} \cmidrule(lr){6-9} \cmidrule(lr){10-13}
 & \multicolumn{2}{c}{TR} & \multicolumn{2}{c}{IR} & \multicolumn{2}{c}{TR} & \multicolumn{2}{c}{IR} & \multicolumn{2}{c}{TR} & \multicolumn{2}{c}{IR} \\
 & R@1 & R@10 & R@1 & R@10 & R@1 & R@10 & R@1 & R@10 & R@1 & R@10 & R@1 & R@10 \\ \toprule[1pt]
1.0 & 52.52 & 17.58 & 60.24 & 33.32 & 56.69 & 21.65 & 65.88 & 39.69 & 63.65 & 31.46 & 72.88 & 45.60 \\
2.0 & 69.33 & 39.84 & 79.22 & 57.63 & 76.56 & 41.97 & 82.64 & 60.47 & 80.40 & 53.81 & 85.81 & 66.77 \\
3.0 & 83.93 & 56.61 & 88.95 & 72.53 & 87.12 & 60.98 & 91.40 & 75.45 & 92.20 & 75.75 & 93.74 & 80.63 \\
4.0 & 89.57 & 70.33 & 93.27 & 82.40 & 92.52 & 74.29 & 95.10 & 84.67 & 96.10 & 86.47 & 96.86 & 89.03 \\
5.0 & \textbf{93.87} & \textbf{78.35} & \textbf{96.33} & \textbf{88.40} & \textbf{94.97} & \textbf{83.54} & \textbf{97.29} & \textbf{90.12} & \textbf{98.21} & \textbf{91.78} & \textbf{98.19} & \textbf{92.79} \\ \toprule[1pt]
\end{tabular}
\end{table*}

\subsubsection{Maximum Number of Replaced Words and Image Perturbation Bound}
Although this work follows previous work \cite{coattack,sga}, setting the perturbation constraint for the text modality (maximum number of replaced words) as $\epsilon_{t}=1$  and for the image modality (image perturbation bound) as $\epsilon_{x}=2/255$, the transferability of a good attack pipeline should show a positive correlation with the perturbation constraint of the corresponding modality. Therefore, we present the changes in transferability under different constraints in \autoref{txt_bound} and \autoref{img_bound}. The results show that as the constraints increase, transferability also increases. Notably, when the perturbation constraint for the image modality is only $\epsilon_{x}=1/255$, the transferability from ALBEF to $\rm CLIP_{ViT}$ SimVLA is higher than that of DRA with $\epsilon_{x}=2/255$, further demonstrating the advantages of SimVLA.

\begin{table*}[!t] 
	\caption{The impact of the image perturbation bound $\epsilon_{x}$ on the attack success rate, using ALBEF as the surrogate model.}
    	\label{img_bound}
	\centering
    \small	
    \setlength{\tabcolsep}{3.1mm}
\begin{tabular}{ccccccccccccc}
\toprule[1pt]
\multirow{3}{*}{$\epsilon_{x}$} & \multicolumn{4}{c}{$\rm CLIP_{ViT}$} & \multicolumn{4}{c}{$\rm CLIP_{CNN}$} & \multicolumn{4}{c}{TCL} \\ \cmidrule(lr){2-5} \cmidrule(lr){6-9} \cmidrule(lr){10-13}
 & \multicolumn{2}{c}{TR} & \multicolumn{2}{c}{IR} & \multicolumn{2}{c}{TR} & \multicolumn{2}{c}{IR} & \multicolumn{2}{c}{TR} & \multicolumn{2}{c}{IR} \\
 & R@1 & R@10 & R@1 & R@10 & R@1 & R@10 & R@1 & R@10 & R@1 & R@10 & R@1 & R@10 \\ \toprule[1pt]
1/255 & 47.61 & 14.84 & 57.96 & 31.90 & 53.84 & 19.11 & 64.11 & 37.48 & 46.26 & 13.73 & 58.88 & 31.78 \\
2/255 & 52.52 & 17.58 & 60.24 & 33.32 & 56.69 & 21.65 & 65.88 & 39.69 & 63.65 & 31.46 & 72.88 & 45.60 \\
4/255 & 55.95 & 21.75 & 64.76 & 37.70 & 60.98 & 26.32 & 70.26 & 43.41 & 74.18 & 44.09 & 78.98 & 55.61 \\
8/255 & 63.31 & 29.78 & 70.04 & 44.13 & 68.34 & 34.76 & 75.26 & 50.20 & 82.82 & 57.72 & 86.26 & 64.87 \\
16/255 & \textbf{72.15} & \textbf{41.67} & \textbf{77.16} & \textbf{54.21} & \textbf{76.20} & \textbf{46.44} & \textbf{82.44} & \textbf{58.81} & \textbf{91.04} & \textbf{76.35} & \textbf{91.93} & \textbf{77.74} \\ \toprule[1pt]
\end{tabular}
\end{table*}

\begin{table*}[!t] 
	\caption{Influence of modality selection in cross-modal word identification, using ALBEF as the surrogate model. 
		}
	\label{modality}
	\centering
	\small
	\setlength{\tabcolsep}{2.2mm}
\begin{tabular}{ccccccccccccc}
\toprule[1pt]
\multirow{3}{*}{Modality} & \multicolumn{4}{c}{$\rm CLIP_{ViT}$} & \multicolumn{4}{c}{$\rm CLIP_{CNN}$} & \multicolumn{4}{c}{TCL} \\ \cmidrule(lr){2-5} \cmidrule(lr){6-9} \cmidrule(lr){10-13}
 & \multicolumn{2}{c}{TR} & \multicolumn{2}{c}{IR} & \multicolumn{2}{c}{TR} & \multicolumn{2}{c}{IR} & \multicolumn{2}{c}{TR} & \multicolumn{2}{c}{IR} \\
 & R@1 & R@10 & R@1 & R@10 & R@1 & R@10 & R@1 & R@10 & R@1 & R@10 & R@1 & R@10 \\ \toprule[1pt]
Text & 42.94 & 13.92 & 56.02 & 29.42 & 50.18 & 14.43 & 61.76 & 34.39 & 61.85 & 28.66 & 71.12 & 42.59 \\
Average Text \& Image & 50.80 & \textbf{17.58} & 58.41 & 32.62 & 55.81 & 20.63 & 65.24 & \textbf{39.69} & \textbf{64.23} & 31.17 & 73.26 & \textbf{45.41} \\
SimVLA (Image) & \textbf{52.52} & \textbf{17.58} & \textbf{60.24} & \textbf{33.32} & \textbf{56.69} & \textbf{21.65} & \textbf{65.88} & \textbf{39.69} & 63.65 & \textbf{31.46} & \textbf{72.88} & 45.60 \\ \toprule[1pt]
\end{tabular}
\end{table*}

\begin{table*}[!t] 
	\caption{Influence of transformation selection in cross-modal word identification, using ALBEF as the surrogate model. 
		}
	\label{transformation}
	\centering
	\small
	\setlength{\tabcolsep}{2.6mm}
\begin{tabular}{ccccccccccccc}
\toprule[1pt]
\multirow{3}{*}{Transformation} & \multicolumn{4}{c}{$\rm CLIP_{ViT}$} & \multicolumn{4}{c}{$\rm CLIP_{CNN}$} & \multicolumn{4}{c}{TCL} \\ \cmidrule(lr){2-5} \cmidrule(lr){6-9} \cmidrule(lr){10-13}
 & \multicolumn{2}{c}{TR} & \multicolumn{2}{c}{IR} & \multicolumn{2}{c}{TR} & \multicolumn{2}{c}{IR} & \multicolumn{2}{c}{TR} & \multicolumn{2}{c}{IR} \\
 & R@1 & R@10 & R@1 & R@10 & R@1 & R@10 & R@1 & R@10 & R@1 & R@10 & R@1 & R@10 \\ \toprule[1pt]
Sacle & 49.20 & 16.46 & \textbf{60.31} & 33.73 & 57.18 & 21.54 & 65.75 & 38.70 & 63.22 & 30.36 & 72.36 & 45.27 \\
Rotation & 51.17 & \textbf{17.78} & 60.15 & 34.19 & 56.56 & \textbf{22.36} & \textbf{66.78} & 38.90 & 63.54 & 30.86 & 72.40 & 44.66 \\
Cropping & 49.33 & 17.48 & 60.12 & \textbf{34.47} & \textbf{58.53} & 20.93 & 66.01 & 39.03 & \textbf{64.17} & 30.26 & 72.38 & \textbf{45.92} \\
Random Noise & \textbf{52.52} & 17.58 & 60.24 & 33.32 & 56.69 & 21.65 & 65.88 & \textbf{39.69} & 63.65 & \textbf{31.46} & \textbf{72.88} & 45.60 \\ \toprule[1pt]
\end{tabular}
\end{table*}

\begin{table*}[!t] 
	\caption{Influence of stage deletion towards different three-stage baselines. The computation time of the entire dataset is given in parentheses.
		}
	\label{tab:abl_stage3}
	\centering
	\small
	\setlength{\tabcolsep}{1.9mm}
    \setlength{\extrarowheight}{1pt} 
\begin{tabular}{ccccccccccccc}
\toprule[1pt]
\multirow{3}{*}{Attack} & \multicolumn{4}{c}{$\rm CLIP_{ViT}$} & \multicolumn{4}{c}{$\rm CLIP_{CNN}$} & \multicolumn{4}{c}{TCL} \\ \cmidrule(lr){2-5} \cmidrule(lr){6-9} \cmidrule(lr){10-13} 
 & \multicolumn{2}{c}{TR} & \multicolumn{2}{c}{IR} & \multicolumn{2}{c}{TR} & \multicolumn{2}{c}{IR} & \multicolumn{2}{c}{TR} & \multicolumn{2}{c}{IR} \\
 & R@1 & R@10 & R@1 & R@10 & R@1 & R@10 & R@1 & R@10 & R@1 & R@10 & R@1 & R@10 \\ \toprule[1pt]
SGA (3213s) & 32.39 & 8.64 & 44.07 & 21.00 & 35.12 & 10.50 & 46.79 & 22.97 & 45.84 & 16.13 & 55.43 & 27.82 \\
SGA w/o Stage 3 (\textbf{3034s})  & \textbf{35.71} & \textbf{8.94} & \textbf{48.23} & \textbf{22.92} & \textbf{42.82} & \textbf{11.99} & \textbf{54.80} & \textbf{27.82} & \textbf{46.68} & \textbf{15.53} & \textbf{57.02} & \textbf{28.94} \\ \hline
DRA (6238s)& \textbf{36.69} & 8.84 & 47.97 & 22.76 & 39.59 & 11.02 & 49.78 & 24.92 & \textbf{48.16} & \textbf{17.84} & 56.40 & 29.41 \\
DRA w/o Stage 3 (\textbf{6013s}) & 36.56 & \textbf{9.15} & \textbf{48.39} & \textbf{22.90} & \textbf{42.70} & \textbf{12.20} & \textbf{54.70} & \textbf{27.78} & 47.10 & 16.13 & \textbf{57.05} & \textbf{29.57} \\ \hline
SA-AET (8040s)& 37.81 & 10.96 & \textbf{50.48} & \textbf{25.51} & \textbf{45.40} & 13.21 & 54.57 & 28.37 & \textbf{54.69} & \textbf{24.55} & 64.19 & 36.74 \\
SA-AET w/o Stage 3 (\textbf{7755s}) & \textbf{38.18} & \textbf{11.06} & 49.19 & 24.03 & 44.17 & \textbf{13.82} & \textbf{54.99} & \textbf{28.66} & 53.00 & 23.95 & \textbf{64.62} & \textbf{36.76} \\ \toprule[1pt]
\end{tabular}
\end{table*}

\begin{figure*}[!t] 
\centering
\includegraphics[width=1.4\columnwidth]{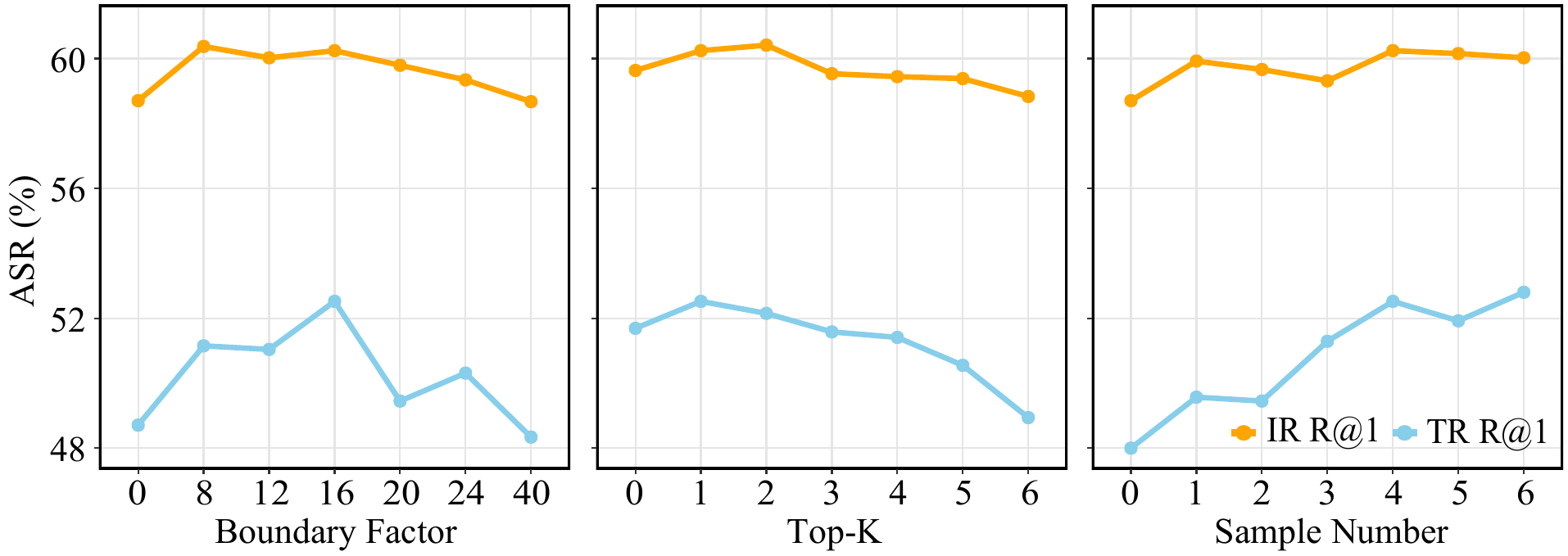}
\caption{
	Parameter sensitivity analysis.
	\textbf{Left:} the boundary factor of sampled images $\beta$ in cross-modal word identification (Stage 1),
	\textbf{Middle:} the number of removed most irrelevant words Top-$K$ in text semantic abstraction (Stage 2), and
	\textbf{Right:} the sample number $N_I$ in cross-modal word identification (Stage 1), and optimization of adversarial image (Stage 2).  
    Here, the surrogate (target) model is ALBEF ($\rm CLIP_{ViT}$).
    }
\label{parameters}
\end{figure*}

\subsubsection{Modality Selection in Cross-Modal Word Identification}
In cross-modal word identification, we propose using only information from the visual modality, specifically the center of the image embedding from a group of sampled images, to guide word identification. 
Here, we provide a detailed comparison between using only textual information and using both visual and textual information by averaging the text and image embeddings produced by the surrogate model's textual and visual encoders.
\autoref{modality} shows that using only textual information, as in previous works \cite{sga,dra}, results 
in the worst transferability. Once visual modality information is incorporated, the transferability is significantly improved, highlighting the importance of selecting the correct modality to guide the generation of adversarial text.

\subsubsection{Transformation Selection in Cross-Modal Word Identification}
In cross-modal word identification, we add random noise to obtain the sampled images. Here, we also compare it with other transformations. 
As shown in \autoref{transformation} the performance of other transformations is similar to the random noise. Considering that our goal is to emphasize the cross-modal guidance design, i.e., the proposed cross-modal word identification, rather than the particular choice of transformations, we therefore adopt simple random noise as the final transformation.

\subsubsection{Stage Rebundancy in Baselines}
Table \autoref{abl} shows that removing the third stage improves transferability in most cases. 
This suggests that Stage~3 may introduce redundant interactions and even hurt generalization across models. We further examine whether this finding also applies to other three-stage baselines. As reported in Table \autoref{tab:abl_stage3}, removing the third stage consistently benefits the three baselines on average. This finding suggests that removing Stage~3 can improve efficiency without loss of attack effectiveness.

\subsubsection{Parameter Sensitivity} We analyze the sensitivity of three main parameters in Stage 1 and Stage 2.
For the boundary factor of sampled images $\beta$ in cross-modal word identification (Stage 1), \autoref{parameters} (left) shows that a moderate value yields the best results.  
This is because a larger value makes the sampled image far from the input image while a smaller value cannot provide enough auxiliary information.
For the number of removed most irrelevant words Top-$K$ in text semantic abstraction (Stage 2), \autoref{parameters} (middle) shows that removing 1-2 words yields relatively strong attack results, while removing too many words leads to a decline in performance due to excessive information loss.
For the sampled number $N_I$ used in cross-modal word identification (Stage 1) and adversarial image optimization (Stage 2), \autoref{parameters} (right) shows that sampling from 4 to 6 images yields relatively strong performance. Considering that previous attacks such as SGA and DRA typically sample four images, we therefore adopt $N_I = 4$ in our final setting.
Additionally, TR results in \autoref{parameters} tend to show higher variances than the IR results, because TR averages over fewer queries, its ASR estimate has higher variance from a statistical perspective.

\section{Conclusion and Discussion}

In this paper, we revisit the common three-stage attack pipeline of current transferable attacks on vision-language pre-training models (VLPMs). 
We identify one problem per attack and propose the Simple Vision-Language Attack (SimVLA) pipeline to address them.
Extensive experiments across various downstream tasks, datasets, and models confirm the superiority of SimVLA in both transferability and computational efficiency.
The significant improvement achieved by our SimVLA exactly highlights the necessity of thoroughly analyzing and leveraging domain knowledge and the fundamental characteristics of the task, whereas blindly adopting intricate operations can be ineffective. 
Besides, the simplicity of our SimVLA also facilitates potential extensions for better performance.
For future work, for image representation, it is worth exploring more advanced strategies beyond only using noisy images to compute image embedding centers.
Additionally, for word replacement, future work may consider determining the replacement based on gradients beyond directly using words generated by BERT.

The proposed SimVLA can be applied in several practical scenarios. First, it can serve as a diagnostic and stress-testing tool for evaluating the robustness of VLPMs and MLLMs before deployment, including black-box or API-based systems, by identifying weaknesses in text--image alignment that may lead to incorrect or unreliable outputs.
Second, insights gained from our SimVLA may inform the design of more robust training objectives and defense mechanisms, contributing to the development of safer and more reliable VLPMs and MLLMs.


\section{Acknowledgments}
This work was supported by the New Generation Artificial Intelligence-National Science and Technology Major Project (2025ZD0123305), the Fundamental and Interdisciplinary Disciplines Breakthrough Plan of the Ministry of Education of China (JYB2025XDXM114), and the National Natural Science Foundation of China (62406240, U2441240, T2341003, 62521002, U24B20185, 62376210).

\appendix
\section{Overlap of Deleted Words}
\label{ap:ssec:delete_stat}
Here, we provide the statistical result of the overlapping deleted words across different surrogate models under the Top-1 settings. As shown in \autoref{ap:tab:delete_stat}, across different surrogate models, the text semantic abstraction can identify similar words, validating its ability to explore the irrelevant words toward the paired image.

\begin{table}[t] 
	\caption{Proportion that our text semantic abstraction can select the same Top-1 word using any two different models across the entire dataset. The proportion is much higher than the random selection (9.37\%).}
	\label{ap:tab:delete_stat}
\centering
\setlength{\tabcolsep}{3.0mm}
\small
\begin{tabular}{ccccc}
\toprule[1pt]
Model & ALBEF & TCL & $\rm CLIP_{ViT}$ & $\rm CLIP_{CNN}$  \\ \toprule[1pt]
ALBEF & - & 55.72\% & 35.02\% & 41.47\% \\
TCL & 55.72\% & - & 36.56\% & 40.06\% \\
$\rm CLIP_{ViT}$ & 55.72\% & 36.56\% & - & 47.16\% \\
$\rm CLIP_{CNN}$ & 41.47\% & 40.06\% & 47.16\% & -  \\ \toprule[1pt]
\end{tabular}
\end{table}

\section{Details about MIM-PGIA}
\label{MIM-PGIA}

Results in the \autoref{2ta} reveal that excessively decreasing the cosine similarity at the Stage 3 (the second text attack stage) will cause overfitting to the surrogate model. 
This motivates us to question whether similar overfitting may occur in the Stage 2 (the image attack stage) when attacking VLPMs, given that image attacks typically require more iterations and are consequently more prone to overfitting to the surrogate model. 
Therefore, we further incorporate the recent theory of flatness into the image attack stage to mitigate potential overfitting.

\begin{figure}[t] 
	\centering
	
	\includegraphics[width=0.7\columnwidth]{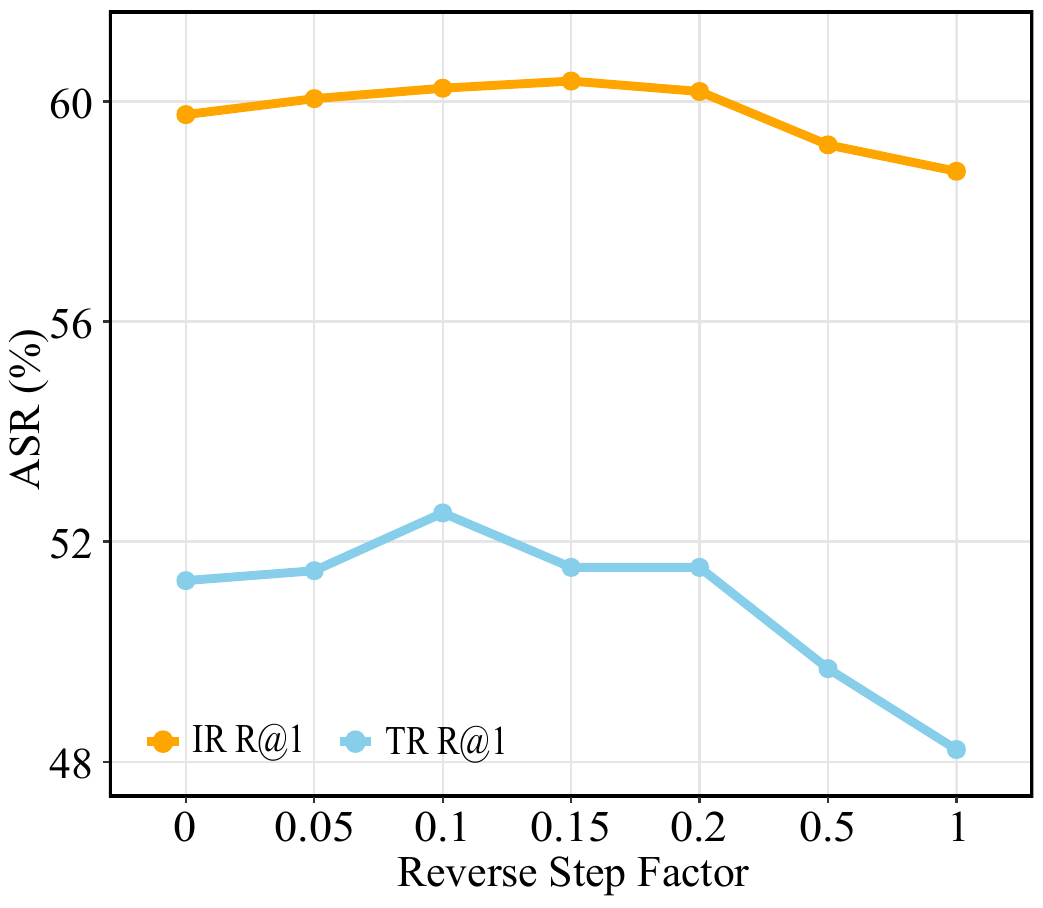}
	\caption{
		Parameter sensitivity analysis of the reverse step factor $\gamma$. 
		Here, attack success rate on R@1 is adopted because this can clearly reflect the change trend. Surrogate model is ALBEF and target model is $\rm CLIP_{ViT}$.}
	\label{rf}
\end{figure}

Assuming each image is paired with $N_T$ texts, the loss function is given as:
\begin{equation}
	\label{one_image_loss_app}
	L(x) = - \sum\limits_{j = 1}^{{N_T}} {\frac{{{f_I}(x) \cdot {f_T}({{({t_j})_{adv}^R}})}}{{{{\left\| {{f_I}(x)} \right\|}_2} \cdot {{\left\| {{f_T}({{({t_j})_{adv}^R}})} \right\|}_2}}}},
\end{equation} 
Once the text attack stage is complete, the text embedding $ f_T(({t_j})_{adv}^R) $ is fixed. 
Therefore, for simplicity, we denote the loss function $L(x)$ solely in terms of the input image $ x $.

RAP \cite{rap} proposes to optimize the following max-min bi-level optimization problem to obtain a flat maximum adversarial loss for the adversarial image:
\begin{equation}
	\label{bi_image_loss}
	{x_{adv}} = \mathop {\max }\limits_{x \in B(x,\epsilon_{x})} \mathop {\min }\limits_{x' \in B(x,\xi_{x} )} L(x),
\end{equation} 
where $\xi_{x}$ is the search radius of the inner optimization. 
This optimization problem involves first locating a position around the input with a relatively low loss, then generating an adversarial image based on that position.

However, directly solving the max-min bi-level optimization problem needs plenty of iteration, which is time-consuming. We hence introduce a recently proposed momentum-based Previous Gradient Inversion Approximation (MIM-PGIA) \cite{ncs} to solve Equation (\ref{bi_image_loss}). 

First, we find the inner solution using PGIA with one iteration ($i=1,2,...,N_{I}$):
\begin{equation}
	\label{pgia}
	\left\{ \begin{array}{l}
		{x_{t - 1,i}} = {x_{t - 1}} + {\delta _i}\\
		{{x'}_{t,i}} = {x_{t,i}} - \gamma  \cdot \frac{{{g_{t - 2,i}} - {g_{t - 1,i}}}}{{{{\left\| {{g_{t - 2,i}} - {g_{t - 1,i}}} \right\|}_2}}}
	\end{array} \right.,
\end{equation}
where ${{x'}_{t,i}}$ is the solution of the inner optimization, ${x_{t - 1}}$ is the adversarial image at the $t-1$ ($t=1,...,T$) iteration of the outer optimization, $\delta _i$ is the random noise, $\gamma$ is the reverse step factor and ${g_{t - 1,i}} = {\nabla _{{{x'}_{t - 1,i}}}}L({{x'}_{t - 1,i}})$.

Then, the outer optimization is solved by MIM \cite{mi}:
\begin{equation}
	\label{outer}
	\left\{ \begin{array}{l}
		{m_{t + 1}} = {m_t} + \frac{1}{{{N_I}}}\sum\limits_{i = 1}^{{N_I}} {\frac{{{\nabla _{{{x'}_{t - 1,i}}}}L({{x'}_{t - 1,i}})}}{{{{\left\| {{\nabla _{{{x'}_{t - 1,i}}}}L({{x'}_{t - 1,i}})} \right\|}_1}}}} \\
		{x_{t + 1}} = {x_{t,i}} + \alpha  \cdot sign({m_{t + 1}})
	\end{array} \right.,
\end{equation}
where $m_t$ is the momentum term and $\alpha$ is the step size.

\autoref{rf} indicates that the optimal value of the reverse step factor $\gamma$ is around 0.1. 
A small value cannot effectively move away from the current high-loss position to a flatter region, while a large value will undermine the adversarial effects accumulated by previous iterations.

\section{MLLM Evaluation Details}
\label{mllm_eval_details} 

{
The prompt template used for the MLLM evaluation is shown below. For simplicity, minor differences between the open-source and closed-source MLLM APIs are omitted.
}

\begin{tcolorbox}[
    enhanced,
    breakable,
    colback=gray!3,
    colframe=black!55,
    boxrule=0.4pt,
    arc=1pt,
    left=2pt,
    right=2pt,
    top=1.5pt,
    bottom=1.5pt,
    boxsep=0.5pt,
    before skip=2pt,
    after skip=2pt,
]
You are a system that judges whether a text description matches an image.
Analyze the image and the input text, and decide if the text description
is present in the image.

\textbf{Matching Rules (be slightly flexible):}

\begin{itemize}
    \item Focus on main objects and key attributes; small differences, such as extra colors or minor details, are acceptable.
    \item If the image mostly fits the text (e.g., an orange-and-white  hat for ``an orange hat''), treat it as a match. Use \texttt{[mismatch]} only when a main object or key attribute is clearly absent or contradicted.
\end{itemize}

\textbf{Instructions:}

\begin{enumerate}
    \item If the text matches the image under the above rules, reply with \texttt{[match]} and briefly explain why it matches.
    \item If the text does not match the image, reply with
    \texttt{[mismatch]} and briefly explain why.
\end{enumerate}

\textbf{Output Examples:}

Example 1: ``\texttt{[match]} The text describes the image accurately.''

Example 2: ``\texttt{[mismatch]} The text describes a dog, but it is not
present in the image.''

\textbf{Input:}
\texttt{<image>} and \texttt{<adversarial text>}.
\end{tcolorbox}

{ All adversarial examples are generated with the random seed fixed to 42, consistent with the settings used for baselines (such as SGA, DRA, SA-AET), and our SimVLA, and are kept unchanged during the MLLM evaluation. For the local models, deterministic decoding is used: sampling is disabled for Qwen2.5-VL, while LLaVA uses a temperature of 0 and top-$k$ of 1. 
For the closed-source APIs, an explicit reproducibility seed is not exposed by the evaluation interface. We therefore fix the model snapshots, input pairs, prompt, and request parameters for all methods. Specifically, the evaluated model snapshots are ``gpt-4.1-nano-2025-04-14,'' ``gpt-4.1-mini-2025-04-14,'' ``gpt-5-nano-2025-08-07,'' ``gpt-5-mini-2025-08-07,'' ``gpt-5.4-2026-03-05,'' and ``gpt-5.5-2026-04-23.'' Considering the API cost, each pair is evaluated once under this fixed protocol.

The outputs are evaluated automatically rather than manually. For each query that successfully returns a non-empty model response, we perform string-based matching on the output label. A query is considered valid only if a model response is successfully obtained after the predefined retry procedure. An attack is counted as successful if the valid response contains ``[mismatch],'' indicating that the target MLLM no longer recognizes the adversarial text and image as a matching pair; otherwise, it is counted as unsuccessful. The accompanying explanations are not used for scoring. The ASR is calculated as the number of successful attacks divided by the total number of valid queries.
}

{
    \small
    \bibliographystyle{ieeenat_fullname}
    \bibliography{reference}
}
\end{document}